\documentclass[11pt]{article}
\usepackage[]{acl}

\providecommand{\low}[1]{\cellcolor{red!15}#1}
\usepackage{multirow}
\usepackage{amsmath}
\usepackage{amssymb}
\usepackage{subcaption}
\usepackage{xspace}
\usepackage{booktabs,colortbl}
\newcommand{\method}{\textsc{Perspective-Driven Inference}\xspace}
\newcommand{\nburnin}{n_{\mathrm{burnin}}}
\newcommand{\nhuman}{n_{\mathrm{human}}}
\newcommand{\E}{\mathbb{E}}

\usepackage{longtable}  
\usepackage{array}
\usepackage{float}

\usepackage{times}
\usepackage{latexsym}
\newcommand\blfootnote[1]{%
  \begingroup
  \renewcommand\thefootnote{}\footnote{#1}%
  \addtocounter{footnote}{-1}%
  \endgroup
}

\usepackage[T1]{fontenc}

\usepackage[utf8]{inputenc}

\usepackage{microtype}

\usepackage{inconsolata}

\usepackage{cuted}

\usepackage{graphicx}
\usepackage{tabularx}
\usepackage{booktabs}

\title{Multi-Perspective LLM Annotations for Valid Analyses in Subjective Tasks}
\newcommand\coauth{$^\star$}
\author{Navya Mehrotra\coauth \\
  Johns Hopkins University \\
  \texttt{nmehrot2@jhu.edu} \\\And
  Adam Visokay\coauth \\
  University of California, Berkeley \\
  \texttt{avisokay@berkeley.edu} \\\And
 Kristina Gligorić\\
  Johns Hopkins University \\
  \texttt{gligoric@jhu.edu} \\
  }

\begin{document}
\maketitle
\begin{abstract}
Large language models are increasingly used to annotate texts, but their outputs reflect some human perspectives better than others. Existing methods for correcting LLM annotation error assume a single ground truth. However, this assumption fails in subjective tasks where disagreement across demographic groups is meaningful. Here we introduce \method, a method that treats the distribution of annotations across groups as the quantity of interest, and estimates it using a small human annotation budget. We contribute an adaptive sampling strategy that concentrates human annotation effort on groups where LLM proxies are least accurate. We evaluate on politeness and offensiveness rating tasks, showing targeted improvements for harder-to-model demographic groups relative to uniform sampling baselines, while maintaining coverage.
\end{abstract}
\blfootnote{\coauth Equal contribution.}

\section{Introduction}

LLMs are promising tools for reducing annotation burden in computational social science (CSS) \cite{ziems2024can, manning2024automated}. Yet, LLM annotations can be wrong \cite{dillion2023can}, encode particular positionalities \cite{santy-etal-2023-nlpositionality}, and reproduce stereotypes \cite{shrawgi2024uncovering}. 

Methods like Prediction-Powered Inference \cite{doi:10.1126/science.adi6000} and Design-Based Supervised Learning \cite{egami2023using} offer a repair~\cite{hullman2026human}, addressing these limitations by combining potentially inaccurate LLM proxies with a small set of ground truth human labels, yielding valid downstream inference even when the LLM proxy is imperfect.

\begin{figure*}[ht!]
    \centering
    \includegraphics[width=\linewidth]{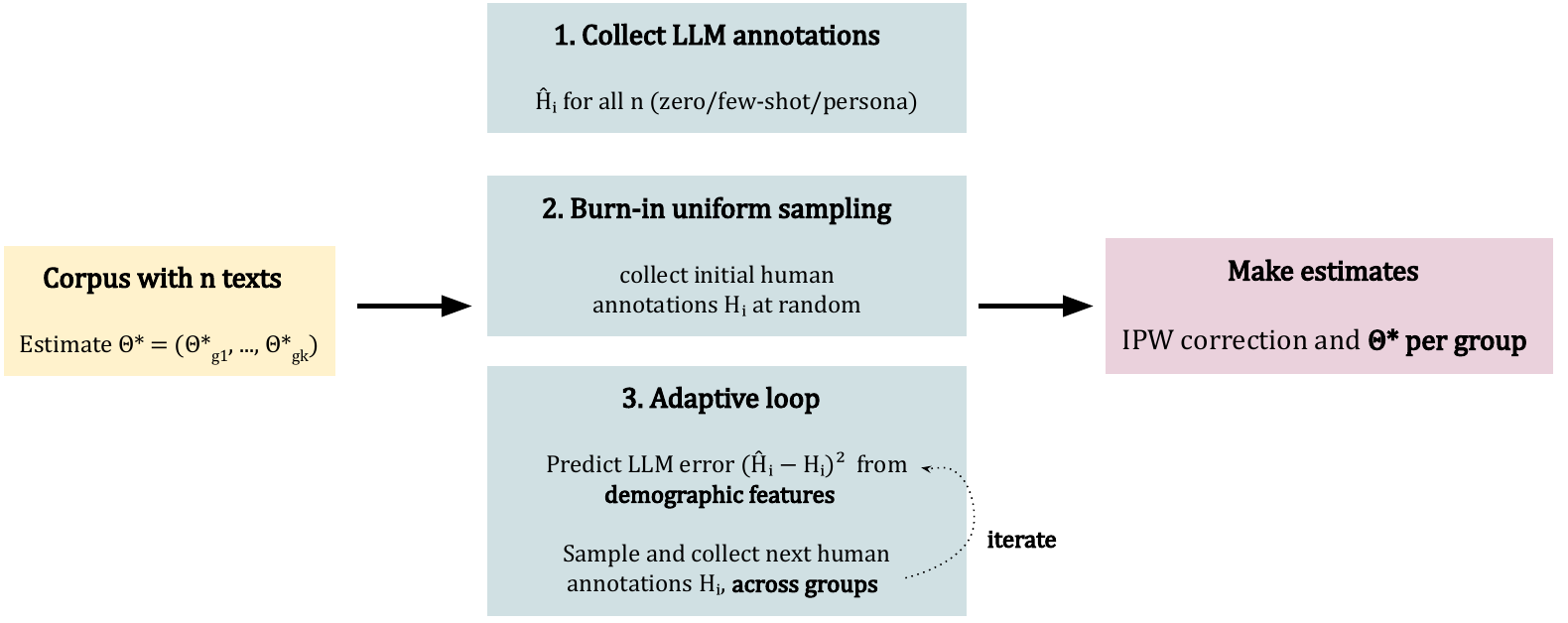}
    \caption{\textbf{Overview of \method.} Starting from a corpus of $n$ texts, we collect LLM annotations, initialize human annotation via uniform sampling, and then enter an adaptive loop that predicts LLM error from demographic features, sampling human annotations across groups. The core idea is to prioritize human annotation for texts where LLMs poorly represent the target demographic group. Once the annotation budget is exhausted, we apply inverse probability weighting (IPW) to produce group-level estimates  $\theta^*$.}
    \label{fig:method}
\end{figure*}

However, these repair methods assume a ground truth exists. In many CSS settings, ground truth does not exist. Reasonable people disagree about whether language is toxic, empathetic, or polite, and that disagreement is not an error to be corrected, but \emph{a signal to be preserved}. Collapsing annotator disagreement into a majority label discards the very variation that matters for understanding how people use and perceive language.

A common method to address this is persona prompting~\cite{truong2025persona}: conditioning LLMs on demographic attributes to simulate the perspectives of specific groups. However, persona prompting does not always improve performance of large language models~\cite{zheng2024helpful,sun2025sociodemographic}: it can exaggerate group differences \cite{cheng2023compost}, misrepresent minority viewpoints \cite{wang2025large}, and in many settings performs no better than a generic prompt \cite{zheng2024helpful, sun2025sociodemographic,xu2026consensusperspectivistmodelingevaluation}.

Here, we introduce \method, a framework for valid corpus inference in subjective annotation tasks where no single ground truth exists (illustrated in Figure~\ref{fig:method}). Our goal is not to train LLMs to make more accurate predictions, but to leverage their annotations to conduct robust downstream statistical inference that faithfully retains annotator disagreement. 

Our contributions are:
\begin{enumerate}
\item \textbf{Framework.} We formalize multi-perspective corpus inference as estimation of a vector estimand (one entry per demographic group), rather than a single scalar, and show how standard PPI-based correction extends to this setting.
\item \textbf{Sampling method.} We develop a sampling strategy that concentrates the human annotation budget on demographic groups for which LLM annotations are predicted to be less reliable, with the goal of improving inference for underrepresented or poorly modeled perspectives.
\item \textbf{Evaluation.} We evaluate \method on three datasets: politeness ratings, offensiveness ratings, and synthetic binary annotations, finding improved accuracy for targeted demographic groups, while maintaining coverage.

\end{enumerate}

We release code and data at \url{https://github.com/JHU-CSS/PDI} to support tasks where retaining differences in annotator perspectives is desired.

\section{Background}

\paragraph{Annotators' background} A growing body of NLP literature views annotation disagreement not as noise, but as a valuable signal~\cite{uma2021learning,hovy2015demographic}. Prior work has shown considerable heterogeneity in annotator preferences on subjective tasks, ranging from toxicity~\cite{sap2022annotators} and sexism detection~\cite{tahaei-bergler-2024-analysis}, to theory of mind modeling \cite{fan2025somi} and counterspeech generation~\cite{mun2024counterspeakers}, with demographic axes like gender~\cite{biester2022analyzing,chen2025feels} and race~\cite{lee2024people} shifting annotations systematically~\cite{al-kuwatly-etal-2020-identifying}. Modeling this disagreement explicitly, rather than collapsing it into majority vote, has become a concern in NLP \cite{fleisig2023majority,xu2026consensusperspectivistmodelingevaluation,xu2025modeling}. Recent efforts push toward characterizing how demographic factors shape human preferences in LLM outputs~\cite{movva2025s,chen2025steer}, and using that understanding to build systems that represent multiple perspectives~\cite{sorensenposition, feng2024modular,rottger2022two}. What is different about our work is that we ask: how can we efficiently leverage limited budgets of human input, given that models do not capture different perspectives equally well~\cite{sen2025missing,santurkar2023whose}? In doing so, our goal is not to use LLMs to make better predictions, but to use LLM annotations as inputs to downstream analyses, while preserving diverse perspectives on subjective tasks. 

\paragraph{Valid inference with proxy annotations} Corpus-level inference from text has roots in literary analysis \cite{yavuz-2019-analyses} and causal inference in NLP \cite{feder2022causal}. Early work formalized the task of inferring document class prevalence under uncertainty \cite{keith2018uncertainty}. A separate thread examined whether annotations themselves are valid~\cite{valentino2021natural,ruder2025assessing}, and how annotation error and genuine label variation can be disentangled \cite{weber-genzel-etal-2024-varierr, baledent2022validity}. More recent work in CSS turns to downstream inference under annotation imperfection, using LLM outputs as surrogates for human labels and developing statistically rigorous methods for correction \cite{wang2020methods,egami2023using,angelopoulos2023ppi++}. We contribute to this line of work by complicating the assumption that a ground truth exists at all: in subjective tasks, disagreement is constitutive, and we show how these repair methods apply in settings where no single correct label can be defined, and the target estimand is a vector, rather than a scalar.

\paragraph{Adaptive sampling} Active learning methods reduce annotation cost by selecting the most informative instances for labeling \cite{baldridge-osborne-2004-active, arora-etal-2009-estimating}, with established tools to support these workflows~\cite{lin-etal-2019-alpacatag, tsvigun-etal-2022-altoolbox, nghiem-ananiadou-2018-aplenty}. Work in the multi-annotator setting specifically targets instances where annotators are likely to disagree \cite{baumler-etal-2023-examples}, and recent methods allocate annotation effort between humans and LLMs based on uncertainty \cite{li2023coannotating, rouzegar-makrehchi-2024-enhancing, zhang-etal-2023-data}. Most related to our approach is sampling based on LLM annotation confidence~\cite{gligoric-etal-2025-unconfident}, and stratified PPI~\cite{fisch2024stratified}, which improves inference efficiency by partitioning samples into strata. In contrast, our stratification is along \emph{perspectives} rather than strata with data subsets: we sample human annotators to cover the distribution of viewpoints, not to reduce label uncertainty around a single ground truth.

\section{Methods}

\subsection{Problem setup}

We have a corpus of $n$ independent and identically distributed (i.i.d.) text instances $T_1,\dots,T_n$. We consider a \emph{perspective-aware} setup: each instance is annotated by multiple annotators, each coming from a particular perspective. Specifically, each observation is a (text, annotator) pair $(T_i, d_i)$, where $d_i \in \mathcal{D}$ encodes the demographic profile of the annotator (e.g., gender, race, age, education level).

The human annotation $H_i$ therefore captures the judgment of an annotator with demographics $d_i$ on text $T_i$ (i.e., index $i$ runs over (text, annotator) pairs). Note that we focus on demographic features as a proof of concept. In principle, $d_i$ features could include any axis along which human perspectives vary, informed by theoretical background knowledge, or exploratory analyses.

We partition annotators into $K$ demographic groups $\mathcal{G} = \{g_1, \dots, g_K\}$. We wish to estimate a \emph{vector} of group-specific quantities
\[
\theta^* = (\theta^*_{g_1}, \dots, \theta^*_{g_K}),
\]
where each $\theta^*_{g_k} = \E[H_i \mid d_i \in g_k]$ is the expected annotation value within demographic group $g_k$, i.e., the group mean. Note that, beyond the group means, the framework applies for standard statistics such as group medians or regression coefficients~\cite{angelopoulos2023ppi++}.

For instance, $\theta^*_{g_k}$ can be the fraction of texts rated as polite by annotators in group $g_k$, or their mean perceived offensiveness rating. Estimating $\theta^*$ separately per group reveals how a quantity of interest differs across demographic perspectives.

In addition to human annotations, we have access to LLM annotations $\hat H_i$ for all $n$ instances. These serve as a cheap but potentially inaccurate proxy for human judgments. We define the \emph{LLM error} for instance $i$ as $(\hat H_i - H_i)^2$.

Our goal is to estimate $\theta^*$ with a valid confidence interval at a pre-specified level $(1-\alpha)$, using only $\nhuman \ll n$ human annotations in total. Two challenges make this goal difficult: (i) rare demographic groups are likely to receive too few annotations under uniform sampling, yielding imprecise group-level estimates; and (ii) LLM annotations may be systematically biased for certain demographic perspectives.

\subsection{Perspective-driven inference}

To overcome these challenges, we combine LLM annotations with \emph{adaptive} sampling of human annotations, to produce an unbiased estimate $\hat\theta_{\mathrm{PDI}}$ of each $\theta^*_{g_k}$, together with a valid confidence interval. 

The adaptive design concentrates the human annotation budget on demographic groups for which LLM annotations are \emph{least reliable}, improving precision for poorly-modeled groups.

\paragraph{Adaptive sampling rule}
We collect a human annotation $H_i$ for observation $(T_i, d_i)$ with probability $\pi_i$. Let $\xi_i = \mathbf{1}\{H_i \text{ collected}\}$ denote the corresponding indicator. Motivated by active inference \cite{zrnic2024active,gligoric-etal-2025-unconfident}, we set
\[
\pi_i \propto \widehat{\texttt{err}}_i(d_i),
\]
normalized so that $\sum_{i=1}^n \pi_i = \nhuman$. The function $\widehat{\texttt{err}}_i$ is a black-box predictor of the LLM squared error trained on annotator demographics. Crucially, $H_i$ is unknown before annotation, so $\pi_i$ is approximated by learning a predictor of the LLM error from annotator demographic features.

As annotations are collected, we use $\{(d_j, (\hat H_j - H_j)^2)\}_{j < i,\, \xi_j = 1}$ as feature--label pairs to fit $\widehat{\texttt{err}}_i$ via a black-box regressor (gradient-boosted regression tree, XGBoost). That is, we train a model to predict the squared LLM error from the annotator's demographic profile, and then allocate human annotations to groups with high predicted error.

To bootstrap the sampling rule, we first collect $\nburnin$ annotations uniformly at random (referred to as the \emph{burn-in} phase), which are part of the total annotation budget. We then proceed in $B$ batches. Within each batch $b$, we normalize the predicted errors into a discrete probability vector, and smooth it with the uniform distribution over the batch to prevent zero-probability allocations. We refine $\widehat{\texttt{err}}_i$ after each batch using all annotations collected so far.

\paragraph{Perspective-driven estimator}

After collecting human annotations according to $\pi_i$, we estimate each group mean $\theta^*_{g_k}$ via an inverse-probability-weighted (IPW) rectified estimator (Appendix, Sec.~\ref{equation}). We form a valid $(1-\alpha)$-level confidence interval for each $\theta^*_{g_k}$ via bootstrap~\cite{zrnic2024active}.

\subsection{Baselines}

\paragraph{LLM only (zero-shot)} This baseline prompts the LLM with the task description and text $T_i$ directly, with no examples, and treats the resulting annotation as ground truth. No human annotation budget is required, but the estimate may be biased if the LLM misrepresents certain demographic perspectives, or is inaccurate, depending on which demographics LLMs default to during annotation~\cite{schafer-etal-2025-demographics}.

\paragraph{LLM only (few-shot)} This baseline augments the zero-shot prompt with a small number of randomly chosen labeled examples. The few-shot examples may improve accuracy relative to zero-shot prompting, but demographic bias can persist.

\paragraph{LLM only (persona prompt)} This baseline instructs the LLM to adopt a specific demographic persona before annotating. Persona prompting is designed to steer the LLM toward a particular demographic perspective, but fidelity to that perspective is not guaranteed.

\paragraph{PPI (uniform sampling)} This baseline uses the same rectified estimator as PDI but collects human annotations \emph{uniformly at random} rather than adaptively, i.e., $\pi_i = \nhuman/n$ for all $i$. It is a special case of $\hat\theta_{\mathrm{PDI}}$ with uniform $\pi_i$ and optimally chosen $\lambda$. Since uniform sampling ignores group membership, minority demographic groups may receive few annotations. This baseline corresponds to the estimator of \citet{angelopoulos2023ppi++}.

For both PPI and PDI, for each model, we use the LLM variant (zero-shot, few-shot, or persona) that achieves the highest coverage for the LLM-only estimator, selected separately for each model. We note that we are aiming to achieve targeted sub-improvements, rather than perform universally superior compared to PPI.

\subsection{Evaluation}

\paragraph{Datasets} 

We evaluate on both real-world and synthetic data. For real-world data, we use two annotation tasks from POPQUORN \citep{pei-jurgens-2023-annotator}, a large-scale dataset that collects demographically diverse human annotations via Prolific. POPQUORN contains 45,000 annotations from 1,484 annotators across multiple tasks.  The texts come from three sources: the Enron email corpus (emails), Reddit (social media comments), and SQuAD (Wikipedia-based reading comprehension questions posed by crowdworkers). We use the politeness rating and offensiveness rating subsets, taking 1,000 unique annotations from each task. For additional robustness experiments, we construct a synthetic dataset of $N=10,000$ items drawn from two demographic groups with selected label rates and group-specific LLM error rates (see Appendix, Section~\ref{sec:sythetic} for details).

\paragraph{Annotation tasks}

In the politeness task, annotators rate how polite a statement would be in an email from a colleague on a Likert scale from 1.0 (not polite at all) to 5.0 (very polite). In the offensiveness task, annotators rate how offensive a statement would be as a Reddit comment on the same scale. Synthetic annotations are binary.

\paragraph{Inference tasks}

Our goal is to estimate the vector $\theta^* = (\theta^*_{g_1}, \dots, \theta^*_{g_K})$, where each $\theta^*_{g_k}$ is the mean annotation rating within demographic group $g_k$. For the politeness task, $\theta^*_{g_k}$ is the mean perceived politeness rating among annotators in group $g_k$; for the offensiveness task, it is the mean perceived offensiveness rating. We stratify separately by gender, age, and education, yielding $K$ groups per demographic axis, used as features in the black box regressor. 

\paragraph{Descriptive statistics}

\begin{figure*}[t]
\centering
\includegraphics[width = \linewidth]{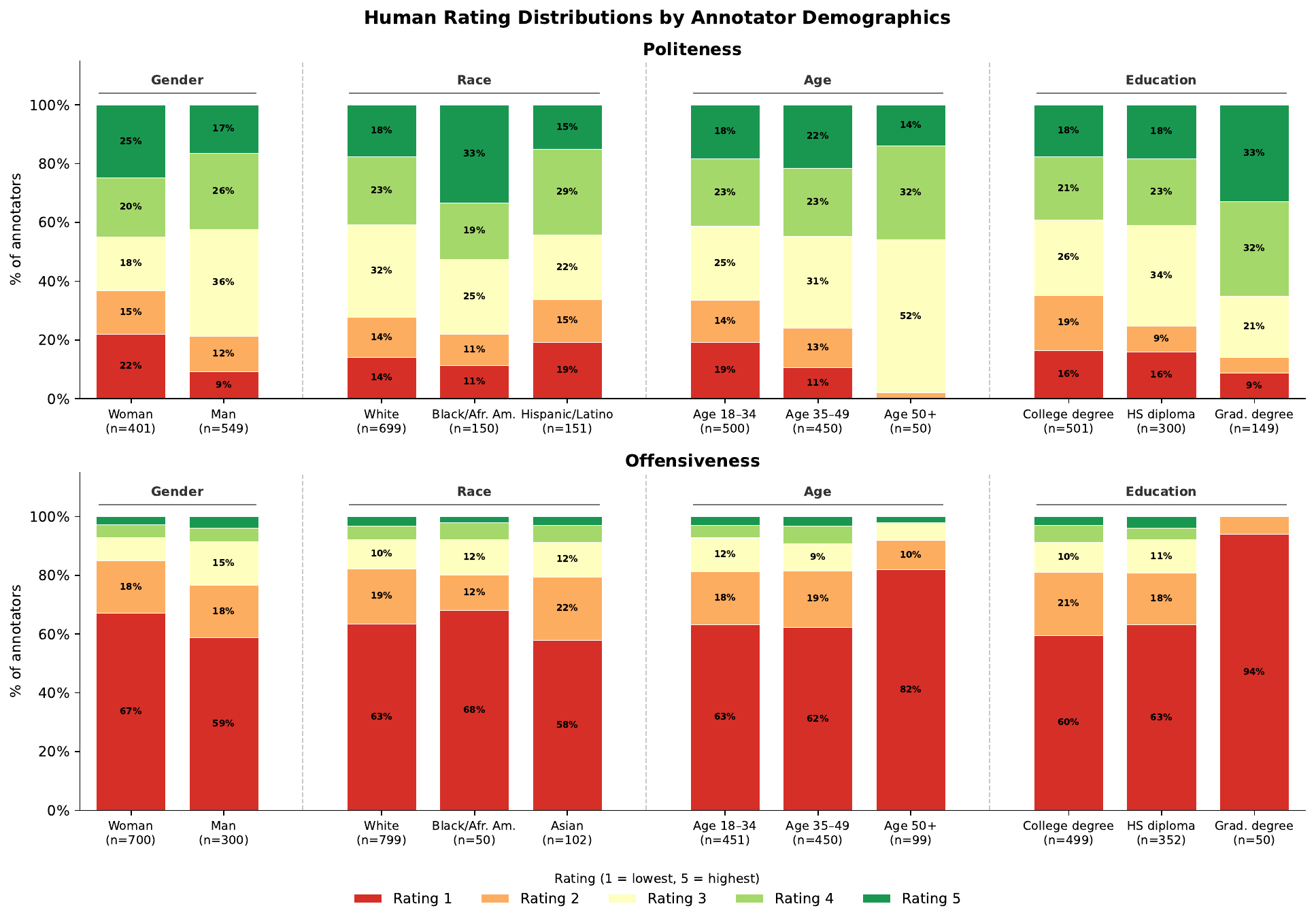}
\caption{\textbf{Annotation distributions vary across demographic groups.} Human ratings for politeness (top) and offensiveness (bottom) broken down by annotator demographics. Variation across groups motivates estimating a vector of group-specific means rather than a single aggregate.}
\label{fig:annotation-distributions}
\end{figure*}

Annotation ratings vary significantly across demographic groups in both real-world datasets, motivating the multi-perspective estimation. We verify this using chi-squared tests on ratings. In the offensiveness dataset, we find significant variation by gender ($\chi^2(4) = 13.305$, $p = 0.0099$), with men rating statements as more offensive on average than women (1.77 vs.\ 1.58), as well as by age ($\chi^2(28) = 90.261$, $p < 0.001$) and education ($\chi^2(16) = 49.837$, $p < 0.001$). The politeness dataset shows the same pattern such that ratings differ significantly by gender, age, and education. Since the differences between age groups are most pronounced in both politeness and offensiveness ratings (Figure~\ref{fig:annotation-distributions}), we focus on age strata in the evaluation. For illustration, Tables~\ref{tab:annotation-examples-off} and ~\ref{tab:annotation-examples-pol} list example texts, along with annotations. For example, ``Women's studies is dumb though, stop defending your degree?'' was generally seen as more offensive by younger annotators: those aged 18--24 and 25--29 rated it 5, while most annotators aged 50 and above rated it 1, with one 50--54-year-old exception who rated it 5. A similar pattern appears for ``Still not as scary as autism.'', which received two ratings of 5 from 18--24-year-olds, whereas most annotators aged 45 and above rated it 1, although one 54--59-year-old rated it 5.

\paragraph{Models}

We evaluate across eight LLMs spanning a range of model families and scales: GPT-5.2, GPT-OSS-120B, Claude Opus 4.6, Claude Sonnet 4.6, Claude Haiku 4.5, Gemini 3.1 Pro Preview, Llama 3 8B Instruct, and Mistral Large 3. Each model is prompted to produce offensiveness and politeness ratings on the same statements presented to human annotators, using identical task descriptions and rating scales. Full prompt texts and prompting parameters such as temperature are listed in Appendix, Section~\ref{app:prompts}. 

\paragraph{Evaluation metrics}

We evaluate methods in terms of \emph{coverage} and \emph{mean absolute error}.

\noindent \underline{\textit{Coverage.}} Coverage is the empirical rate at which the produced confidence intervals contain the true $\theta^*_{g_k}$. Since $\theta^*_{g_k}$ is unknown, we approximate it by the group mean computed on the full dataset. For PDI and PPI, coverage is provably valid by construction for large $n$~\cite{angelopoulos2023ppi++}; we report it as a sanity check and to evaluate the LLM-only baselines, which have no coverage guarantees.

\noindent \underline{\textit{Delta.}} For each estimator, we measure how far its point estimate deviates from the human-rated group mean $\theta^*_{g_k}$, approximated by the full-dataset group mean. Specifically, we report the average absolute error in percentage points, averaged across demographic groups. Lower ${\Delta}$ indicates that the method's estimates are closer to ground truth on average. This metric captures estimation accuracy directly in the units of the quantity of interest (percentage points). Coverage asks whether the interval is in the right place, while delta asks whether the estimate itself is. Metrics are reported across 20 trials.

\begin{table*}[t]
\centering
\small
\setlength{\tabcolsep}{5pt}
\begin{tabular}{lcccccccc}
\toprule
& \multicolumn{2}{c}{\textbf{Avg (Age)}} 
& \multicolumn{2}{c}{\textbf{Age 18--34}} 
& \multicolumn{2}{c}{\textbf{Age 35--49}} 
& \multicolumn{2}{c}{\cellcolor{gray!20}\textbf{Age 50+}} \\
\cmidrule(lr){2-3}\cmidrule(lr){4-5}\cmidrule(lr){6-7}\cmidrule(lr){8-9}
\textbf{Method} & Cov. & $\Delta$ & Cov. & $\Delta$ & Cov. & $\Delta$ & \cellcolor{gray!20}Cov. & \cellcolor{gray!20}$\Delta$ \\
\midrule
GPT-5.2 (zero shot)   & \cellcolor{red!15}88.3 & 7.48 & \cellcolor{green!20}95.0 & 4.10 & \cellcolor{red!15}80.0 &  6.88 & \cellcolor{gray!20}\cellcolor{green!20}90.0 & \cellcolor{gray!20}11.46 \\
GPT-5.2 (few shot)    & \cellcolor{green!20}91.7 & 7.52 & \cellcolor{green!20}100.0 & 2.45 & \cellcolor{green!20}100.0 &  3.81 & \cellcolor{gray!20}\cellcolor{red!15}75.0 & \cellcolor{gray!20}16.31 \\
GPT-5.2 (persona)     & \cellcolor{green!20}93.3 & 7.02 & \cellcolor{green!20}100.0 & 3.58 & \cellcolor{green!20}95.0 &  4.60 & \cellcolor{gray!20}\cellcolor{red!15}85.0 & \cellcolor{gray!20}12.87 \\
PPI                   & \cellcolor{green!20}95.0 & 6.35 & \cellcolor{green!20}100.0 & 2.93 & \cellcolor{green!20}100.0 &  2.49 & \cellcolor{gray!20}\cellcolor{red!15}85.0 & \cellcolor{gray!20}13.63 \\
PDI                   & \cellcolor{green!20}93.3 & 7.10 & \cellcolor{green!20}90.0 & 5.51 & \cellcolor{green!20}95.0 &  4.58 & \cellcolor{gray!20}\cellcolor{green!20}{95.0} & \cellcolor{gray!20}{11.23} \\
\bottomrule
\end{tabular}
\caption{\textbf{Politeness.} Coverage (Cov.) and delta ($\Delta$) by age group on the politeness task (GPT-5.2). Values below 90\% coverage are \colorbox{red!15}{highlighted}.}
\label{tab:politeness-age}
\end{table*}

\begin{table}[t]
\centering
\small
\setlength{\tabcolsep}{6pt}
\begin{tabular}{lrr}
\toprule
\textbf{Group} & \textbf{PDI samples} & \textbf{Uniform samples} \\
\midrule
\multicolumn{3}{l}{\textit{Gender}} \\
\quad Woman             &  78.7 &  85.1 \\
\quad Man               & 121.3 & 114.9 \\
\midrule
\multicolumn{3}{l}{\textit{Race}} \\
\quad White             & 135.8 & 140.2 \\
\quad Black/Afr.\ Am.  &  34.3 &  28.8 \\
\quad Hispanic/Latino  &  29.9 &  31.1 \\
\midrule
\multicolumn{3}{l}{\textit{Age}} \\
\quad Age 18--34        & 105.0 & 106.8 \\
\quad Age 35--49        &  82.9 &  84.0 \\
\quad \cellcolor{gray!20}Age 50+  & \cellcolor{gray!20}{12.2} & \cellcolor{gray!20}9.2 \\
\midrule
\multicolumn{3}{l}{\textit{Education}} \\
\quad College degree    &  83.8 &  94.8 \\
\quad HS diploma        &  73.2 &  63.6 \\
\quad Grad.\ degree     &  32.8 &  31.9 \\
\bottomrule
\end{tabular}
\caption{\textbf{Politeness.} Average number of human annotations allocated per group by PDI and PPI, averaged across 20 trials (total budget $200$ per trial). PDI allocates 33\% more annotations to Age 50+ than PPI (12.2 vs.\ 9.2), reflecting error-driven upsampling toward the hardest subgroup.}
\label{tab:sampling-counts-pol}
\end{table}

\section{Results}

\subsection{Politeness} In the main text we focus on GPT-5.2 (results with the other models are reported in the Appendix, Table~\ref{tab:appendix_llm_eval}). Across prompting variants, the few-shot configuration yields the smallest mean prediction error on the politeness task. The subpopulation with the largest error under the best-performing variant is \textbf{Age 50+}, with a few-shot error of 0.426. 

\paragraph{Coverage} First, we find that coverage consistently falls below 90\% for several LLM-only methods on older age groups (Table~\ref{tab:politeness-age}). Zero-shot GPT-5.2 misses the target for Age 35--49 (coverage = 0.800). For Age 50+, both the few-shot variant (coverage = 0.750) and the persona variant (coverage = 0.850) fall short. In contrast, PDI maintains coverage at or above 90\% across all three age groups, reaching 0.950 for Age 50+.

\paragraph{Delta} Averaged across age groups, the LLM-only methods are comparable to PPI and non-rectified baselines: zero-shot and few-shot GPT-5.2 produce average deltas of 7.48 and 7.52 percentage points, respectively, versus 6.35 for PPI and 7.10 for PDI. Again, the critical divergence appears for \textbf{Age 50+}: The few-shot variant is inaccurate for this group (coverage = 0.750), producing the worst delta at 16.31\%. In contrast, PDI achieves the lowest delta for Age 50+ at 11.23\%, outperforming PPI (13.63\%) and all three LLM-only variants.

\paragraph{Upsampling lift} The lower delta for Age 50+ under PDI reflects error-driven upsampling. Because the error regressor accurately identifies this group as high-error, PDI directs a disproportionate share of the annotation budget there (Table~\ref{tab:sampling-counts-pol}). Averaged across 20 trials, PPI allocates 9.2 annotations to Age 50+ under uniform sampling; PDI allocates 12.2, a 33\% increase. This confirms that the predicted error signal is useful: the regressor correctly identifies where annotation effort matters most, and the resulting upsampling translates into improved coverage and accuracy for the most challenging subpopulation. 

Crucially, this reallocation does not compromise overall validity---averaged across all three age groups, PDI maintains 0.933 coverage.

\begin{table*}[t]
\centering
\small
\setlength{\tabcolsep}{5pt}
\begin{tabular}{lcccccccc}
\toprule
& \multicolumn{2}{c}{\textbf{Avg (Age)}} 
& \multicolumn{2}{c}{\textbf{Age 18--34}} 
& \multicolumn{2}{c}{\textbf{Age 35--49}} 
& \multicolumn{2}{c}{\cellcolor{gray!20}\textbf{Age 50+}} \\
\cmidrule(lr){2-3}\cmidrule(lr){4-5}\cmidrule(lr){6-7}\cmidrule(lr){8-9}
\textbf{Method} & Cov. & $\Delta$ & Cov.& $\Delta$ & Cov. & $\Delta$ & \cellcolor{gray!20}Cov. & \cellcolor{gray!20}$\Delta$ \\
\midrule
GPT-5.2 (zero shot)   & \cellcolor{red!15}11.7 & 20.11 & \cellcolor{red!15}15.0 & 14.05 & \cellcolor{red!15}5.0  & 18.38 & \cellcolor{gray!20}\cellcolor{red!15}15.0  & \cellcolor{gray!20}27.90 \\
GPT-5.2 (few shot)    & \cellcolor{red!15}53.3 & 14.21 & \cellcolor{red!15}65.0 &  8.01 & \cellcolor{red!15}60.0 & 10.06 & \cellcolor{gray!20}\cellcolor{red!15}35.0  & \cellcolor{gray!20}24.54 \\
GPT-5.2 (persona)     & \cellcolor{red!15}13.3 & 19.68 & \cellcolor{red!15}25.0 & 13.12 & \cellcolor{red!15}0.0  & 17.50 & \cellcolor{gray!20}\cellcolor{red!15}15.0  & \cellcolor{gray!20}28.40 \\
PPI                   & \cellcolor{green!20}95.0 &  3.86 & \cellcolor{green!20}100.0 &  2.38 & \cellcolor{green!20}95.0 &  3.57 & \cellcolor{gray!20}\cellcolor{green!20}90.0  & \cellcolor{gray!20}5.64 \\
PDI                   & \cellcolor{green!20}95.0 &  6.04 & \cellcolor{green!20}95.0  &  7.31 & \cellcolor{green!20}95.0 &  5.57 & \cellcolor{gray!20}\cellcolor{green!20}{95.0}  & \cellcolor{gray!20}{5.24} \\
\bottomrule
\end{tabular}
\caption{\textbf{Offensiveness.} Coverage (Cov.) and delta ($\Delta$) by age group on the offensiveness task (GPT-5.2). Values below the 90\% coverage are \colorbox{red!15}{highlighted}.}
\label{tab:offensiveness-age}
\end{table*}

\begin{table}[t]
\centering
\small
\setlength{\tabcolsep}{6pt}
\begin{tabular}{lrr}
\toprule
\textbf{Group} & \textbf{PDI samples} & \textbf{PPI samples} \\
\midrule
\multicolumn{3}{l}{\textit{Gender}} \\
\quad Woman             & 152.3 & 143.7 \\
\quad Man               &  47.6 &  56.4 \\
\midrule
\multicolumn{3}{l}{\textit{Race}} \\
\quad White             & 164.7 & 162.9 \\
\quad Black/Afr.\ Am.  &   6.8 &   9.0 \\
\quad Asian             &  17.4 &  20.2 \\
\midrule
\multicolumn{3}{l}{\textit{Age}} \\
\quad Age 18--34        &  85.1 &  87.0 \\
\quad Age 35--49        &  89.2 &  91.4 \\
\quad \cellcolor{gray!20}Age 50+  & \cellcolor{gray!20}{25.8} & \cellcolor{gray!20}21.6 \\
\midrule
\multicolumn{3}{l}{\textit{Education}} \\
\quad College degree    & 102.4 & 103.2 \\
\quad HS diploma        &  67.5 &  68.5 \\
\quad Grad.\ degree     &  12.2 &  11.4 \\
\bottomrule
\end{tabular}
\caption{Average number of human annotations allocated per group by PDI and PPI, averaged across 20 trials (total budget $200$ per trial, offensiveness task). PDI allocates 19\% more annotations to Age 50+ than PPI (25.8 vs.\ 21.6), reflecting error-driven upsampling toward the hardest subgroup.}
\label{tab:offensiveness-sampling-counts}
\end{table}

\subsection{Offensiveness}
For the offensiveness task, the persona variant results in the smallest mean prediction error for GPT-5.2. As with politeness, age is the demographic axis with the greatest variation in performance, and Age 50+ again emerges as the hardest subgroup (Table~\ref{tab:offensiveness-age}).

\paragraph{Coverage} The coverage for offensiveness is substantially worse than for politeness. All three LLM-only methods fail: averaged across age groups, zero-shot GPT-5.2 achieves only 11.7\% coverage, the persona variant 13.3\%, and few-shot variant only 53.3\% (substantially below the 90\% coverage across every age group). For age 35--49, the persona variant achieves 0\% coverage. In contrast, both PPI and PDI maintain 95.0\% average coverage across all three age groups, including Age 50+.

\paragraph{Delta} The LLM-only delta values reflect the same breakdown: zero-shot and persona produce average deltas above 19 percentage points, and even the best LLM-only variant (few-shot, 14.21\%) falls far short of PPI (3.86\%) and PDI (6.04\%). For Age 50+, all three LLM-only variants exceed 24\% delta, with persona reaching 28.40\%. PDI achieves the lowest delta for Age 50+ at 5.24\%, outperforming PPI (5.64\%), while also maintaining coverage.

\paragraph{Upsampling lift} As with politeness, the PDI performance for Age 50+ is supported by error-driven upsampling (Table~\ref{tab:offensiveness-sampling-counts}). PDI allocates on average 25.8 annotations to Age 50+ across 20 trials, compared to 21.6 under uniform PPI sampling (a 19\% increase). The upsampling lift surpasses PPI while preserving validity: averaged across all three age groups, PDI maintains coverage.

\subsection{Synthetic data}
\label{sec :synthetic}
 
To understand the impact of the human annotation budget, size of the burn-in, group skew, and LLM error, we analyze PDI on synthetic data, under controlled conditions. The simulated data consist of $N = 10{,}000$ items drawn from two demographic groups:
a large group (Group 1, $N_1 = 9{,}000$, 90\%) and a small group (Group 2, $N_2 = 1{,}000$, 10\%).
Each item carries a binary human label $Y_i \in \{0,1\}$ and a binary LLM prediction $\hat{Y}_i$.
LLM predictions are generated by flipping the true label independently with group-specific error rates (see Appendix, Sec.~\ref {sec:sythetic} for details). Overall, we find that adaptive sampling is advantageous over uniform sampling when the budget of human annotations is at least 20\%, as sufficient labels are needed to learn the sampling rule (Figure~\ref{fig:synthetic1}), while the choice of precise burn-in size is not consequential, as long as it is under 50\% of the complete budget (Figure~\ref{fig:synthetic2}). As expected, the stronger the disparity in LLM performance between groups and the more severe the group skew, the more \method deviates from uniform sampling (Figures~\ref{fig:synthetic3} and \ref{fig:synthetic4}). These differences are particularly pronounced for group sizes under 10\%, and for error differences above 20\%, where upsampling is particularly valuable, when compared to uniform sampling.

\subsection{Limitations of persona prompting} Lastly, overall across the experiments, we find that persona prompting is not a reliable method. Across models, it is the highest-variance prompting strategy and frequently the worst (Appendix, Sec.~\ref{app:persona}). Persona prompting sometimes matches few-shot performance, but it can also shift predictions systematically in the wrong direction, suggesting the effect depends heavily on model-specific patterns, rather than on the demographic content of the persona itself. Persona prompting is the most common method for capturing demographic perspectives with LLMs, yet it fails precisely for the groups it is meant to represent, and in several cases leaves coverage at zero. The results confirm that conditioning an LLM on a demographic label is not a substitute for human annotation from that group, and underscore why a correction framework that treats LLM outputs as potentially inadequate proxies (rather than steerable perspectives) can be useful.

\section{Discussion}

Overall, our results show that correcting LLM annotation error can be important when the target estimand is a distribution over perspectives rather than a single ground truth. LLM-only methods fail for specific groups that are poorly modeled by persona prompting (such as Age 50+). PDI addresses this by directing human annotation effort where it matters most. The gains can be substantial: for offensiveness, PDI reduces delta for Age 50+ to 5.24\%, while LLM-only variants exceed 24\%. Coverage holds across all groups. Practitioners should expect the largest gains over uniform sampling when LLM performance disparities between groups are large, and groups are skewed.

The evaluation therefore demonstrates that PDI can be viewed as a targeted subgroup-improvement method (rather than a universally superior alternative to uniform PPI). On aggregate metrics, PDI does not consistently outperform uniform PPI. But PDI is designed for settings where the main concern is not average performance across all groups, but better inference for those groups that are hardest for LLM proxies to capture. Our findings support this interpretation, with PDI showing its clearest benefits on harder-to-model groups, such as Age 50+, while maintaining coverage.

The framework may generalize beyond the tasks studied here. More broadly, corpus inference problems where annotator identity shapes labels are promising candidates for this approach. The demographic axes we use are a proof of concept. In practice, the stratification could follow any theoretically motivated partition, including ideological orientation, professional background, or lived experience~\cite{orlikowski2025beyond}. The key requirement is that the error predictor receives informative features. Instead, where demographic proxies are weak, the adaptive gains will be smaller, and the method degrades to uniform PPI.

The problem we study also appears in three high-stakes settings that have received attention in NLP. In RLHF, preference labels reflect the perspectives of whoever provided feedback, yet models are trained as if those preferences were universal~\cite{movva2025s,chen2025steer}. In LLM-as-judge evaluation, a single model scores outputs according to implicit standards that may not generalize across demographic groups~\cite{dong2024can}. In LLM simulations, the task is often to predict distributions of opinions as self-reported in survey answers~\cite{krsteski2025valid,suh2025language}, and LLM predictions do not represent each respondent equally well. All three settings involve collapsing perspective-dependent judgments into a scalar, discarding variation that is not noise. Extending our framework to these settings is a natural direction, and one with implications for how models are trained or evaluated.

Lastly, several limitations remain. Our evaluation uses two tasks, and performance on tasks with subtler demographic structure is a promising avenue for future work. The burn-in phase requires uniform samples before adaptation begins, which may be inefficient when the human budget is very small. The error predictor is a gradient-boosted tree trained on demographic features alone; richer features, including linguistic properties of the text, could improve predictions and allocate the budget more efficiently. 

\section{Conclusion}
LLMs are widely used as annotation proxies, but their outputs do not represent all demographic perspectives equally. We introduced a framework that treats the distribution of group-specific annotations as the estimand of interest, and allocates a limited human annotation budget adaptively to where LLM error is highest. Across two subjective rating tasks, PDI maintains valid coverage and reduces estimation error for the hardest subpopulations. The framework is task-agnostic and can extend to any setting where annotator identity shapes labels, including RLHF, LLM-as-judge evaluation, and survey simulation. Correcting for LLM bias at the aggregate level is not enough; valid inference requires knowing \emph{whose} perspective the model fails to capture, and directing human effort there.

\section*{Limitations}

First, we stratify by gender, age, and education independently, but these axes correlate in practice. An annotator who is older and identifies as a particular gender occupies an intersection that our per-axis treatment does not capture. Modeling intersectional subgroups directly would require larger annotation budgets and raises challenges for the error predictor when intersectional cells are sparse. We leave joint stratification to future work. Because our predictor targets overall squared error, it does not disentangle reducible error (driven by underrepresentation) from irreducible noise (driven by inherent, intra-group heterogeneity). Consequently, while the adaptive rule correctly allocates budget to minority groups to improve precision, it will also oversample adequately represented but highly heterogeneous groups. While this irreducible noise could theoretically be mitigated by further partitioning the data into more homogeneous subgroups, doing so requires robust domain knowledge and experimentation to guarantee that the newly defined strata capture meaningful perspectives.

Second, our evaluation focuses on zero-shot, few-shot, and persona-prompted LLMs without fine-tuning. PDI is agnostic to how the proxy is constructed, and applies equally to fine-tuned models. A fine-tuned proxy that better tracks demographic variation would reduce the correction needed and likely improve efficiency further; our results represent a conservative setting where the proxy is cheap and off-the-shelf.

Third, we evaluate on two subjective rating tasks from a single dataset. Demographic axes are used as a proof of concept: in principle, any partition along which human perspectives vary systematically (like ideological orientation, professional background, lived experience) can serve as the stratification. Broader evaluation across tasks, domains, and non-demographic perspective axes remains an important direction.

Lastly, the adaptive component is deliberately simple: the error predictor relies only on demographic features, rather than richer textual signals that might improve targeting accuracy. In principle, this likely leaves some performance on the table. The current design is a proof of concept: it shows that even a minimal adaptive strategy can deliver meaningful subgroup gains.

\section*{Ethical considerations}

We caution against using demographic persona prompting as a substitute for actual human input. Conditioning LLMs on demographic attributes can misrepresent and exaggerate group identities~\cite{wang2025large}, and our results show that persona prompting often fails precisely for the groups it is meant to represent. Our framework instead treats LLM outputs as potentially inadequate proxies and uses human annotations to correct for that inadequacy.

Our approach estimates group-level means, not individual-level behavior. A finding that Age 50+ annotators rate statements differently on average should not be interpreted as a claim about how any particular older person reasons or perceives language. We aim to capture diversity in aggregate judgments, and caution against misreading our framework as a tool for stereotyping or as evidence of fixed group-level dispositions.

The demographic categories we use (gender, age, education) are coarse proxies for perspective, not deterministic accounts of it. These categories are themselves socially constructed and unevenly operationalized across datasets. Results should be interpreted in light of how demographic attributes were collected in POPQUORN, and may not transfer to populations where those categories have different meaning. The dataset we leverage is publicly available and does not contain personally identifying information, nor does it uniquely identify individual people.

Our framework requires collecting human annotations from demographically identified participants. This raises concerns about participant privacy, informed consent, and the potential for re-identification from demographic metadata. Any deployment should ensure that annotators are fairly compensated, particularly those from underrepresented groups who may be asked to contribute disproportionately under adaptive sampling.

Finally, adaptive upsampling concentrates annotation effort on groups where LLMs perform worst. In practice, these are often already marginalized groups. While this is a feature of our design, it also means those annotators bear more of the annotation labor. This tradeoff should be made explicit to participants, and compensation should reflect it.

\bibliography{custom}

@inproceedings{shrawgi2024uncovering,
  title={Uncovering stereotypes in large language models: A task complexity-based approach},
  author={Shrawgi, Hari and Rath, Prasanjit and Singhal, Tushar and Dandapat, Sandipan},
  booktitle={Proceedings of the 18th Conference of the European Chapter of the Association for Computational Linguistics (Volume 1: Long Papers)},
  pages={1841--1857},
  year={2024}
}

@inproceedings{orlikowski2025beyond,
  title={Beyond demographics: Fine-tuning large language models to predict individuals’ subjective text perceptions},
  author={Orlikowski, Matthias and Pei, Jiaxin and R{\"o}ttger, Paul and Cimiano, Philipp and Jurgens, David and Hovy, Dirk},
  booktitle={Proceedings of the 63rd Annual Meeting of the Association for Computational Linguistics (Volume 1: Long Papers)},
  pages={2092--2111},
  year={2025}
}

@inproceedings{sun2025sociodemographic,
  title={Sociodemographic prompting is not yet an effective approach for simulating subjective judgments with LLMs},
  author={Sun, Huaman and Pei, Jiaxin and Choi, Minje and Jurgens, David},
  booktitle={Proceedings of the 2025 Conference of the Nations of the Americas Chapter of the Association for Computational Linguistics: Human Language Technologies (Volume 2: Short Papers)},
  pages={845--854},
  year={2025}
}

@article{xu2025modeling,
  title={Modeling annotator disagreement with demographic-aware experts and synthetic perspectives},
  author={Xu, Yinuo and Derricks, Veronica and Earl, Allison and Jurgens, David},
  journal={arXiv preprint arXiv:2508.02853},
  year={2025}
}

@inproceedings{mun2024counterspeakers,
  title={Counterspeakers’ perspectives: Unveiling barriers and {{AI}} needs in the fight against online hate},
  author={Mun, Jimin and Buerger, Cathy and Liang, Jenny T and Garland, Joshua and Sap, Maarten},
  booktitle={Proceedings of the 2024 CHI Conference on Human Factors in Computing Systems},
  pages={1--22},
  year={2024}
}

@article{fan2025somi,
  title={{{SoMi-ToM}}: Evaluating Multi-Perspective Theory of Mind in Embodied Social Interactions},
  author={Fan, Xianzhe and Zhou, Xuhui and Jin, Chuanyang and Nottingham, Kolby and Zhu, Hao and Sap, Maarten},
  journal={arXiv preprint arXiv:2506.23046},
  year={2025}
}

@inproceedings{casola-etal-2024-multipico,
  title={{{MultiPICo}}: Multilingual perspectivist irony corpus},
  author={Casola, Silvia and Frenda, Simona and Lo, Soda Marem and Sezerer, Erhan and Uva, Antonio and Basile, Valerio and Bosco, Cristina and Pedrani, Alessandro and Rubagotti, Chiara and Patti, Viviana and others},
  booktitle={Proceedings of the 62nd Annual Meeting of the Association for Computational Linguistics (Volume 1: Long Papers)},
  pages={16008--16021},
  year={2024}
}

@inproceedings{jang-frassinelli-2024-generalizable,
  title={Generalizable sarcasm detection is just around the corner, of course!},
  author={Jang, Hyewon and Frassinelli, Diego},
  booktitle={Proceedings of the 2024 Conference of the North American Chapter of the Association for Computational Linguistics: Human Language Technologies (Volume 1: Long Papers)},
  pages={4238--4249},
  year={2024}
}

@article{akhtar2021whose,
  title={Whose opinions matter? perspective-aware models to identify opinions of hate speech victims in abusive language detection},
  author={Akhtar, Sohail and Basile, Valerio and Patti, Viviana},
  journal={arXiv preprint arXiv:2106.15896},
  year={2021}
}

@inproceedings{rizzi-etal-2024-soft,
  title={Soft metrics for evaluation with disagreements: an assessment},
  author={Rizzi, Giulia and Leonardelli, Elisa and Poesio, Massimo and Uma, Alexandra and Pavlovic, Maja and Paun, Silviu and Rosso, Paolo and Fersini, Elisabetta},
  booktitle={Proceedings of the 3rd Workshop on Perspectivist Approaches to NLP (NLPerspectives)@ LREC-COLING 2024},
  pages={84--94},
  year={2024}
}

@inproceedings{zheng2024helpful,
  title={When” a helpful assistant” is not really helpful: Personas in system prompts do not improve performances of large language models},
  author={Zheng, Mingqian and Pei, Jiaxin and Logeswaran, Lajanugen and Lee, Moontae and Jurgens, David},
  booktitle={Findings of the Association for Computational Linguistics: EMNLP 2024},
  pages={15126--15154},
  year={2024}
}

@inproceedings{pei-jurgens-2023-annotator,
    title = "When Do Annotator Demographics Matter? Measuring the Influence of Annotator Demographics with the {POPQUORN} Dataset",
    author = "Pei, Jiaxin  and
      Jurgens, David",
    editor = "Prange, Jakob  and
      Friedrich, Annemarie",
    booktitle = "Proceedings of the 17th Linguistic Annotation Workshop (LAW-XVII)",
    month = jul,
    year = "2023",
    address = "Toronto, Canada",
    publisher = "Association for Computational Linguistics",
    url = "https://aclanthology.org/2023.law-1.25/",
    doi = "10.18653/v1/2023.law-1.25",
    pages = "252--265"
}

@article{angelopoulos2023ppi++,
  title={{{PPI++}}: Efficient prediction-powered inference},
  author={Angelopoulos, Anastasios N and Duchi, John C and Zrnic, Tijana},
  journal={arXiv preprint arXiv:2311.01453},
  year={2023}
}

@article{feder2022causal,
  title={Causal inference in natural language processing: Estimation, prediction, interpretation and beyond},
  author={Feder, Amir and Keith, Katherine A and Manzoor, Emaad and Pryzant, Reid and Sridhar, Dhanya and Wood-Doughty, Zach and Eisenstein, Jacob and Grimmer, Justin and Reichart, Roi and Roberts, Margaret E and others},
  journal={Transactions of the Association for Computational Linguistics},
  volume={10},
  pages={1138--1158},
  year={2022},
  publisher={MIT Press One Broadway, 12th Floor, Cambridge, Massachusetts 02142, USA~…}
}

@inproceedings{yavuz-2019-analyses,
    title = "Analyses of Literary Texts by Using Statistical Inference Methods",
    author = "Yavuz, Mehmet Can",
    editor = "Bernardi, Raffaella  and
      Navigli, Roberto  and
      Semeraro, Giovanni",
    booktitle = "Proceedings of the Sixth Italian Conference on Computational Linguistics (CLiC-it 2019)",
    month = nov,
    year = "2019",
    address = "Bari, Italy",
    publisher = "CEUR Workshop Proceedings",
    url = "https://aclanthology.org/2019.clicit-1.61/",
    pages = "410--416",
    ISBN = "979-1-280-13600-8"
}

@article{sen2025missing,
  title={Missing the margins: A systematic literature review on the demographic representativeness of {{LLMs}}},
  author={Sen, Indira and Lutz, Marlene and Rogers, Elisa and Garcia, David and Strohmaier, Markus},
  journal={Findings of the Association for Computational Linguistics: ACL 2025},
  pages={24263--24289},
  year={2025}
}

@inproceedings{feng2024modular,
  title={Modular pluralism: Pluralistic alignment via {{multi-LLMm}} collaboration},
  author={Feng, Shangbin and Sorensen, Taylor and Liu, Yuhan and Fisher, Jillian and Park, Chan Young and Choi, Yejin and Tsvetkov, Yulia},
  booktitle={Proceedings of the 2024 Conference on Empirical Methods in Natural Language Processing},
  pages={4151--4171},
  year={2024}
}

@article{wang2025large,
  title={Large language models that replace human participants can harmfully misportray and flatten identity groups},
  author={Wang, Angelina and Morgenstern, Jamie and Dickerson, John P},
  journal={Nature Machine Intelligence},
  volume={7},
  number={3},
  pages={400--411},
  year={2025},
  publisher={Nature Publishing Group UK London}
}

@inproceedings{krsteski2025valid,
  title={Valid survey simulations with limited human data: The roles of prompting, fine-tuning, and rectification},
  author={Krsteski, Stefan and Russo, Giuseppe and Chang, Serina and West, Robert and Gligori{\'c}, Kristina},
  booktitle={Proceedings of the 64th Annual Meeting of the Association for Computational Linguistics (Volume 1: Long Papers)},
  pages={10887--10906},
  year={2026}
}

@inproceedings{suh2025language,
  title={Language model fine-tuning on scaled survey data for predicting distributions of public opinions},
  author={Suh, Joseph and Jahanparast, Erfan and Moon, Suhong and Kang, Minwoo and Chang, Serina},
  booktitle={Proceedings of the 63rd Annual Meeting of the Association for Computational Linguistics (Volume 1: Long Papers)},
  pages={21147--21170},
  year={2025}
}

@inproceedings{santurkar2023whose,
  title={Whose opinions do language models reflect?},
  author={Santurkar, Shibani and Durmus, Esin and Ladhak, Faisal and Lee, Cinoo and Liang, Percy and Hashimoto, Tatsunori},
  booktitle={International conference on machine learning},
  pages={29971--30004},
  year={2023},
  organization={PMLR}
}

@inproceedings{chen2025steer,
  title={Steer-bench: A benchmark for evaluating the steerability of large language models},
  author={Chen, Kai and He, Zihao and Shi, Taiwei and Lerman, Kristina},
  booktitle={Proceedings of the 2025 Conference on Empirical Methods in Natural Language Processing},
  year={2025}
}

@inproceedings{keith2018uncertainty,
  title={Uncertainty-aware generative models for inferring document class prevalence},
  author={Keith, Katherine and O’Connor, Brendan},
  booktitle={Proceedings of the 2018 Conference on Empirical Methods in Natural Language Processing},
  pages={4575--4585},
  year={2018}
}

@article{hullman2026human,
  title={This human study did not involve human subjects: Validating {{LLM}} simulations as behavioral evidence},
  author={Hullman, Jessica and Broska, David and Sun, Huaman and Shaw, Aaron},
  journal={arXiv preprint arXiv:2602.15785},
  year={2026}
}

@article{dillion2023can,
  title={Can {{AI}} language models replace human participants?},
  author={Dillion, Danica and Tandon, Niket and Gu, Yuling and Gray, Kurt},
  journal={Trends in Cognitive Sciences},
  volume={27},
  number={7},
  pages={597--600},
  year={2023},
  publisher={Elsevier}
}

@article{egami2023using,
  title={Using imperfect surrogates for downstream inference: Design-based supervised learning for social science applications of large language models},
  author={Egami, Naoki and Hinck, Musashi and Stewart, Brandon and Wei, Hanying},
  journal={Advances in Neural Information Processing Systems},
  volume={36},
  pages={68589--68601},
  year={2023}
}

@article{lee2024people,
  title={People who share encounters with racism are silenced online by humans and machines, but a guideline-reframing intervention holds promise},
  author={Lee, Cinoo and Gligori{\'c}, Kristina and Kalluri, Pratyusha Ria and Harrington, Maggie and Durmus, Esin and Sanchez, Kiara L and San, Nay and Tse, Danny and Zhao, Xuan and Hamedani, MarYam G and others},
  journal={Proceedings of the National Academy of Sciences},
  volume={121},
  number={38},
  pages={e2322764121},
  year={2024},
  publisher={National Academy of Sciences}
}

@inproceedings{biester2022analyzing,
  title={Analyzing the effects of annotator gender across NLP tasks},
  author={Biester, Laura and Sharma, Vanita and Kazemi, Ashkan and Deng, Naihao and Wilson, Steven and Mihalcea, Rada},
  booktitle={Proceedings of the 1st Workshop on Perspectivist Approaches to NLP@ LREC2022},
  pages={10--19},
  year={2022}
}

@inproceedings{chen2025feels,
  title={“Feels Feminine to Me”: Understanding Perceived Gendered Style through Human Annotations},
  author={Chen, Hongyu and Falk, Neele and Roth, Michael and Falenska, Agnieszka},
  booktitle={Proceedings of the 2025 Conference on Empirical Methods in Natural Language Processing},
  pages={31447--31468},
  year={2025}
}

@article{movva2025s,
  title={What's In My Human Feedback? Learning Interpretable Descriptions of Preference Data},
  author={Movva, Rajiv and Milli, Smitha and Min, Sewon and Pierson, Emma},
  journal={arXiv preprint arXiv:2510.26202},
  year={2025}
}

@article{sorensenposition,
  title={A roadmap to pluralistic alignment},
  author={Sorensen, Taylor and Moore, Jared and Fisher, Jillian and Gordon, Mitchell and Mireshghallah, Niloofar and Rytting, Christopher Michael and Ye, Andre and Jiang, Liwei and Lu, Ximing and Dziri, Nouha and others},
  journal={arXiv preprint arXiv:2402.05070},
  year={2024}
}

@inproceedings{sap2022annotators,
  title={Annotators with attitudes: How annotator beliefs and identities bias toxic language detection},
  author={Sap, Maarten and Swayamdipta, Swabha and Vianna, Laura and Zhou, Xuhui and Choi, Yejin and Smith, Noah A},
  booktitle={Proceedings of the 2022 conference of the north american chapter of the association for computational linguistics: Human language technologies},
  pages={5884--5906},
  year={2022}
}

@inproceedings{fleisig2023majority,
  title={When the majority is wrong: Modeling annotator disagreement for subjective tasks},
  author={Fleisig, Eve and Abebe, Rediet and Klein, Dan},
  booktitle={Proceedings of the 2023 Conference on Empirical Methods in Natural Language Processing},
  pages={6715--6726},
  year={2023}
}

@inproceedings{schafer-etal-2025-demographics,
    title = "Which Demographics do {{LLM}}s Default to During Annotation?",
    author = {Sch{\"a}fer, Johannes  and
      Combs, Aidan  and
      Bagdon, Christopher  and
      Li, Jiahui  and
      Probol, Nadine  and
      Greschner, Lynn  and
      Papay, Sean  and
      Menchaca Resendiz, Yarik  and
      Velutharambath, Aswathy  and
      Wuehrl, Amelie  and
      Weber, Sabine  and
      Klinger, Roman},
    editor = "Che, Wanxiang  and
      Nabende, Joyce  and
      Shutova, Ekaterina  and
      Pilehvar, Mohammad Taher",
    booktitle = "Proceedings of the 63rd Annual Meeting of the Association for Computational Linguistics (Volume 1: Long Papers)",
    month = jul,
    year = "2025",
    address = "Vienna, Austria",
    publisher = "Association for Computational Linguistics",
    url = "https://aclanthology.org/2025.acl-long.848/",
    doi = "10.18653/v1/2025.acl-long.848",
    pages = "17331--17348",
    ISBN = "979-8-89176-251-0"
}

@inproceedings{al-kuwatly-etal-2020-identifying,
    title = "Identifying and Measuring Annotator Bias Based on Annotators' Demographic Characteristics",
    author = "Al Kuwatly, Hala  and
      Wich, Maximilian  and
      Groh, Georg",
    editor = "Akiwowo, Seyi  and
      Vidgen, Bertie  and
      Prabhakaran, Vinodkumar  and
      Waseem, Zeerak",
    booktitle = "Proceedings of the Fourth Workshop on Online Abuse and Harms",
    month = nov,
    year = "2020",
    address = "Online",
    publisher = "Association for Computational Linguistics",
    url = "https://aclanthology.org/2020.alw-1.21/",
    doi = "10.18653/v1/2020.alw-1.21",
    pages = "184--190"
}

@inproceedings{tahaei-bergler-2024-analysis,
    title = "Analysis of Annotator Demographics in Sexism Detection",
    author = "Tahaei, Narjes  and
      Bergler, Sabine",
    editor = "Fale{\'n}ska, Agnieszka  and
      Basta, Christine  and
      Costa-juss{\`a}, Marta  and
      Goldfarb-Tarrant, Seraphina  and
      Nozza, Debora",
    booktitle = "Proceedings of the 5th Workshop on Gender Bias in Natural Language Processing (GeBNLP)",
    month = aug,
    year = "2024",
    address = "Bangkok, Thailand",
    publisher = "Association for Computational Linguistics",
    url = "https://aclanthology.org/2024.gebnlp-1.24/",
    doi = "10.18653/v1/2024.gebnlp-1.24",
    pages = "376--383"
}

@article{fisch2024stratified,
  title={Stratified prediction-powered inference for effective hybrid evaluation of language models},
  author={Fisch, Adam and Maynez, Joshua and Hofer, R and Dhingra, Bhuwan and Globerson, Amir and Cohen, William W},
  journal={Advances in Neural Information Processing Systems},
  volume={37},
  pages={111489--111514},
  year={2024}
}

@inproceedings{nghiem-ananiadou-2018-aplenty,
    title = "{APL}enty: annotation tool for creating high-quality datasets using active and proactive learning",
    author = "Nghiem, Minh-Quoc  and
      Ananiadou, Sophia",
    editor = "Blanco, Eduardo  and
      Lu, Wei",
    booktitle = "Proceedings of the 2018 Conference on Empirical Methods in Natural Language Processing: System Demonstrations",
    month = nov,
    year = "2018",
    address = "Brussels, Belgium",
    publisher = "Association for Computational Linguistics",
    url = "https://aclanthology.org/D18-2019/",
    doi = "10.18653/v1/D18-2019",
    pages = "108--113"
}

@inproceedings{rouzegar-makrehchi-2024-enhancing,
    title = "Enhancing Text Classification through {{LLM}}-Driven Active Learning and Human Annotation",
    author = "Rouzegar, Hamidreza  and
      Makrehchi, Masoud",
    editor = "Henning, Sophie  and
      Stede, Manfred",
    booktitle = "Proceedings of the 18th Linguistic Annotation Workshop (LAW-XVIII)",
    month = mar,
    year = "2024",
    address = "St. Julians, Malta",
    publisher = "Association for Computational Linguistics",
    url = "https://aclanthology.org/2024.law-1.10/",
    doi = "10.18653/v1/2024.law-1.10",
    pages = "98--111"
}

@inproceedings{li2023coannotating,
  title={Coannotating: Uncertainty-guided work allocation between human and large language models for data annotation},
  author={Li, Minzhi and Shi, Taiwei and Ziems, Caleb and Kan, Min-Yen and Chen, Nancy and Liu, Zhengyuan and Yang, Diyi},
  booktitle={Proceedings of the 2023 Conference on Empirical Methods in Natural Language Processing},
  pages={1487--1505},
  year={2023}
}

@inproceedings{tsvigun-etal-2022-altoolbox,
    title = "{ALT}oolbox: A Set of Tools for Active Learning Annotation of Natural Language Texts",
    author = "Tsvigun, Akim  and
      Sanochkin, Leonid  and
      Larionov, Daniil  and
      Kuzmin, Gleb  and
      Vazhentsev, Artem  and
      Lazichny, Ivan  and
      Khromov, Nikita  and
      Kireev, Danil  and
      Rubashevskii, Aleksandr  and
      Shahmatova, Olga  and
      Dylov, Dmitry V.  and
      Galitskiy, Igor  and
      Shelmanov, Artem",
    editor = "Che, Wanxiang  and
      Shutova, Ekaterina",
    booktitle = "Proceedings of the 2022 Conference on Empirical Methods in Natural Language Processing: System Demonstrations",
    month = dec,
    year = "2022",
    address = "Abu Dhabi, UAE",
    publisher = "Association for Computational Linguistics",
    url = "https://aclanthology.org/2022.emnlp-demos.41/",
    doi = "10.18653/v1/2022.emnlp-demos.41",
    pages = "406--434"
}

@inproceedings{lin-etal-2019-alpacatag,
    title = "{A}lpaca{T}ag: An Active Learning-based Crowd Annotation Framework for Sequence Tagging",
    author = "Lin, Bill Yuchen  and
      Lee, Dong-Ho  and
      Xu, Frank F.  and
      Lan, Ouyu  and
      Ren, Xiang",
    editor = "Costa-juss{\`a}, Marta R.  and
      Alfonseca, Enrique",
    booktitle = "Proceedings of the 57th Annual Meeting of the Association for Computational Linguistics: System Demonstrations",
    month = jul,
    year = "2019",
    address = "Florence, Italy",
    publisher = "Association for Computational Linguistics",
    url = "https://aclanthology.org/P19-3010/",
    doi = "10.18653/v1/P19-3010",
    pages = "58--63"
}

@inproceedings{arora-etal-2009-estimating,
    title = "Estimating Annotation Cost for Active Learning in a Multi-Annotator Environment",
    author = "Arora, Shilpa  and
      Nyberg, Eric  and
      Ros{\'e}, Carolyn P.",
    editor = "Ringger, Eric  and
      Haertel, Robbie  and
      Tomanek, Katrin",
    booktitle = "Proceedings of the {NAACL} {HLT} 2009 Workshop on Active Learning for Natural Language Processing",
    month = jun,
    year = "2009",
    address = "Boulder, Colorado",
    publisher = "Association for Computational Linguistics",
    url = "https://aclanthology.org/W09-1903/",
    pages = "18--26"
}

@inproceedings{baumler-etal-2023-examples,
    title = "Which Examples Should be Multiply Annotated? Active Learning When Annotators May Disagree",
    author = "Baumler, Connor  and
      Sotnikova, Anna  and
      Daum{\'e} III, Hal",
    editor = "Rogers, Anna  and
      Boyd-Graber, Jordan  and
      Okazaki, Naoaki",
    booktitle = "Findings of the Association for Computational Linguistics: ACL 2023",
    month = jul,
    year = "2023",
    address = "Toronto, Canada",
    publisher = "Association for Computational Linguistics",
    url = "https://aclanthology.org/2023.findings-acl.658/",
    doi = "10.18653/v1/2023.findings-acl.658",
    pages = "10352--10371"
}

@inproceedings{zrnic2024active,
  title={Active statistical inference},
  author={Zrnic, Tijana and Cand{\`e}s, Emmanuel J},
  booktitle={Proceedings of the 41st International Conference on Machine Learning},
  pages={62993--63010},
  year={2024}
}

@inproceedings{valentino2021natural,
  title={Do natural language explanations represent valid logical arguments? verifying entailment in explainable NLI gold standards},
  author={Valentino, Marco and Pratt-Hartmann, Ian and Freitas, Andr{\'e}},
  booktitle={Proceedings of the 14th International Conference on Computational Semantics (IWCS)},
  pages={76--86},
  year={2021}
}

@techreport{manning2024automated,
  title={Automated social science: Language models as scientist and subjects},
  author={Manning, Benjamin S and Zhu, Kehang and Horton, John J},
  year={2024},
  institution={National Bureau of Economic Research}
}

@article{ziems2024can,
  title={Can large language models transform computational social science?},
  author={Ziems, Caleb and Held, William and Shaikh, Omar and Chen, Jiaao and Zhang, Zhehao and Yang, Diyi},
  journal={Computational Linguistics},
  volume={50},
  number={1},
  pages={237--291},
  year={2024}
}

@inproceedings{cheng2023compost,
  title={{{CoMPosT}}: Characterizing and evaluating caricature in {{LLM}} simulations},
  author={Cheng, Myra and Piccardi, Tiziano and Yang, Diyi},
  booktitle={Proceedings of the 2023 Conference on Empirical Methods in Natural Language Processing},
  pages={10853--10875},
  year={2023}
}

@inproceedings{dong2024can,
  title={Can {{LLM}} be a personalized judge?},
  author={Dong, Yijiang River and Hu, Tiancheng and Collier, Nigel},
  booktitle={Findings of the Association for Computational Linguistics: EMNLP 2024},
  pages={10126--10141},
  year={2024}
}

@inproceedings{baledent2022validity,
  title={Validity, agreement, consensuality and annotated data quality},
  author={Baledent, Ana{\"e}lle and Mathet, Yann and Widl{\"o}cher, Antoine and Couronne, Christophe and Manguin, Jean-Luc},
  booktitle={Proceedings of the Thirteenth Language Resources and Evaluation Conference},
  pages={2940--2948},
  year={2022}
}

@inproceedings{weber-genzel-etal-2024-varierr,
    title = "{V}ari{E}rr {NLI}: Separating Annotation Error from Human Label Variation",
    author = "Weber-Genzel, Leon  and
      Peng, Siyao  and
      De Marneffe, Marie-Catherine  and
      Plank, Barbara",
    editor = "Ku, Lun-Wei  and
      Martins, Andre  and
      Srikumar, Vivek",
    booktitle = "Proceedings of the 62nd Annual Meeting of the Association for Computational Linguistics (Volume 1: Long Papers)",
    month = aug,
    year = "2024",
    address = "Bangkok, Thailand",
    publisher = "Association for Computational Linguistics",
    url = "https://aclanthology.org/2024.acl-long.123/",
    doi = "10.18653/v1/2024.acl-long.123",
    pages = "2256--2269"
}

@inproceedings{ruder2025assessing,
  title={Assessing the reliability and validity of {{GPT-4}} in annotating emotion appraisal ratings},
  author={Ruder, Deniss and Uusberg, Andero and Sirts, Kairit},
  booktitle={Proceedings of the 10th Workshop on Computational Linguistics and Clinical Psychology (CLPsych 2025)},
  pages={1--11},
  year={2025}
}

@inproceedings{baldridge-osborne-2004-active,
    title = "Active Learning and the Total Cost of Annotation",
    author = "Baldridge, Jason  and
      Osborne, Miles",
    editor = "Lin, Dekang  and
      Wu, Dekai",
    booktitle = "Proceedings of the 2004 Conference on Empirical Methods in Natural Language Processing",
    month = jul,
    year = "2004",
    address = "Barcelona, Spain",
    publisher = "Association for Computational Linguistics",
    url = "https://aclanthology.org/W04-3202/",
    pages = "9--16"
}

@inproceedings{zhang-etal-2023-data,
    title = "Data-efficient Active Learning for Structured Prediction with Partial Annotation and Self-Training",
    author = "Zhang, Zhisong  and
      Strubell, Emma  and
      Hovy, Eduard",
    editor = "Bouamor, Houda  and
      Pino, Juan  and
      Bali, Kalika",
    booktitle = "Findings of the Association for Computational Linguistics: EMNLP 2023",
    month = dec,
    year = "2023",
    address = "Singapore",
    publisher = "Association for Computational Linguistics",
    url = "https://aclanthology.org/2023.findings-emnlp.865/",
    doi = "10.18653/v1/2023.findings-emnlp.865",
    pages = "12991--13008"
}

@article{
doi:10.1126/science.adi6000,
author = {Anastasios N. Angelopoulos  and Stephen Bates  and Clara Fannjiang  and Michael I. Jordan  and Tijana Zrnic },
title = {Prediction-powered inference},
journal = {Science},
volume = {382},
number = {6671},
pages = {669-674},
year = {2023},
doi = {10.1126/science.adi6000},
URL = {https://www.science.org/doi/abs/10.1126/science.adi6000},
eprint = {https://www.science.org/doi/pdf/10.1126/science.adi6000}}

@inproceedings{santy-etal-2023-nlpositionality,
    title = "{NLP}ositionality: Characterizing Design Biases of Datasets and Models",
    author = "Santy, Sebastin  and
      Liang, Jenny  and
      Le Bras, Ronan  and
      Reinecke, Katharina  and
      Sap, Maarten",
    editor = "Rogers, Anna  and
      Boyd-Graber, Jordan  and
      Okazaki, Naoaki",
    booktitle = "Proceedings of the 61st Annual Meeting of the Association for Computational Linguistics (Volume 1: Long Papers)",
    month = jul,
    year = "2023",
    address = "Toronto, Canada",
    publisher = "Association for Computational Linguistics",
    url = "https://aclanthology.org/2023.acl-long.505/",
    doi = "10.18653/v1/2023.acl-long.505",
    pages = "9080--9102"
}

@inproceedings{gligoric-etal-2025-unconfident,
    title = "Can Unconfident {{LLM}} Annotations Be Used for Confident Conclusions?",
    author = "Gligoric, Kristina  and
      Zrnic, Tijana  and
      Lee, Cinoo  and
      Candes, Emmanuel  and
      Jurafsky, Dan",
    editor = "Chiruzzo, Luis  and
      Ritter, Alan  and
      Wang, Lu",
    booktitle = "Proceedings of the 2025 Conference of the Nations of the Americas Chapter of the Association for Computational Linguistics: Human Language Technologies (Volume 1: Long Papers)",
    month = apr,
    year = "2025",
    address = "Albuquerque, New Mexico",
    publisher = "Association for Computational Linguistics",
    url = "https://aclanthology.org/2025.naacl-long.179/",
    doi = "10.18653/v1/2025.naacl-long.179",
    pages = "3514--3533",
    ISBN = "979-8-89176-189-6"
}

@inproceedings{truong2025persona,
  title={Persona-augmented benchmarking: Evaluating {{LLMs}} across diverse writing styles},
  author={Truong, Kimberly and Fogliato, Riccardo and Heidari, Hoda and Wu, Steven},
  booktitle={Proceedings of the 2025 Conference on Empirical Methods in Natural Language Processing},
  year={2025}
}

@article{wang2020methods,
  title={Methods for correcting inference based on outcomes predicted by machine learning},
  author={Wang, Siruo and McCormick, Tyler H and Leek, Jeffrey T},
  journal={Proceedings of the National Academy of Sciences},
  volume={117},
  number={48},
  pages={30266--30275},
  year={2020},
  publisher={National Academy of Sciences}
}

@inproceedings{rottger2022two,
  title={Two contrasting data annotation paradigms for subjective {{NLP}} tasks},
  author={R{\"o}ttger, Paul and Vidgen, Bertie and Hovy, Dirk and Pierrehumbert, Janet},
  booktitle={Proceedings of the 2022 conference of the North American chapter of the association for computational linguistics: human language technologies},
  pages={175--190},
  year={2022}
}

@inproceedings{hovy2015demographic,
  title={Demographic factors improve classification performance},
  author={Hovy, Dirk},
  booktitle={Proceedings of the 53rd annual meeting of the association for computational linguistics and the 7th international joint conference on natural language processing (Volume 1: Long papers)},
  pages={752--762},
  year={2015}
}

@article{uma2021learning,
  title={Learning from disagreement: A survey},
  author={Uma, Alexandra N and Fornaciari, Tommaso and Hovy, Dirk and Paun, Silviu and Plank, Barbara and Poesio, Massimo},
  journal={Journal of Artificial Intelligence Research},
  volume={72},
  pages={1385--1470},
  year={2021}
}

@misc{xu2026consensusperspectivistmodelingevaluation,
      title={Beyond Consensus: Perspectivist Modeling and Evaluation of Annotator Disagreement in {{NLP}}}, 
      author={Yinuo Xu and David Jurgens},
      year={2026},
      eprint={2601.09065},
      archivePrefix={arXiv},
      primaryClass={cs.CL},
      url={https://arxiv.org/abs/2601.09065}, 
}

@article{Monroe_Colaresi_Quinn_2017,
  title={Fightin'words: Lexical feature selection and evaluation for identifying the content of political conflict},
  author={Monroe, Burt L and Colaresi, Michael P and Quinn, Kevin M},
  journal={Political Analysis},
  volume={16},
  number={4},
  pages={372--403},
  year={2008},
  publisher={Cambridge University Press}
}

@article{lam2024conceptInduction,
    author = {Lam, Michelle S. and Teoh, Janice and Landay, James and Heer, Jeffrey and Bernstein, Michael S.},
    title = {Concept Induction: Analyzing Unstructured Text with High-Level Concepts Using LLooM},
    year = {2024},
    isbn = {9798400703300},
    publisher = {Association for Computing Machinery},
    address = {New York, NY, USA},
    url = {https://doi.org/10.1145/3613904.3642830},
    doi = {10.1145/3613904.3642830},
    journal= {Proceedings of the 2024 CHI Conference on Human Factors in Computing Systems},
    articleno = {933},
    numpages = {28},
    location = {Honolulu, HI, USA},
    series = {CHI '24}
}

\appendix

\section{Prompting details}\label{app:prompts}

Table~\ref{tab:prompts} lists the complete prompt texts used for annotation. Models were accessed using OpenRouter, using default parameters such as temperature and max token outputs.

\section{PDI Estimator}\label{equation}

For each text $i$ in group $g_k$, we define labeled and unlabeled IPW weights
\[
w^{\mathrm{lab}}_i = \frac{\xi_i}{\pi_i}, \qquad w^{\mathrm{unlab}}_i = \frac{1-\xi_i}{1-\pi_i},
\]
where $\xi_i = \mathbf{1}\{H_i \text{ collected}\}$ and $\pi_i$ is the annotation probability. 

For any quantity $V$, let $\bar{V}^{(w)}_{g_k} = \frac{\sum_{i \in g_k} w_i V_i}{\sum_{i \in g_k} w_i}$ denote the IPW-weighted mean over the group. We estimate $\theta^*_{g_k}$ using PPI++~\cite{angelopoulos2023ppi++} applied within each group, with IPW-weighted means replacing the standard unweighted means to account for selective annotation.

We draw bootstrap resamples and compute $\hat\theta_{\mathrm{PDI},g_k}$ on each resample. The confidence interval is
\[
C_{1-\alpha,g_k} = \bigl[\hat\theta^{(\alpha/2)},\; \hat\theta^{(1-\alpha/2)}\bigr],
\]
where $\hat\theta^{(p)}$ is the $p$-th quantile of the bootstrap distribution of $\hat\theta_{\mathrm{PDI},g_k}$. The validity of the interval holds under adaptive (non-i.i.d.) sampling because of the IPW correction.

\section{LLM-Only coverage is consistently low}

No model--prompting combination achieves the target coverage on either task. The best observed coverage across all 48 combinations is 0.73 (claude-haiku-4.5 few-shot on politeness; gemini-3.1-pro and gpt-oss-120b few-shot on offensiveness). The median coverage is 0.45 for politeness and 0.09 for offensiveness. The root cause is prediction bias rather than sampling variance: with all $n$ predictions available, confidence intervals are narrow, but they are centered on the wrong value. This motivates methods such as PPI\texttt{++} and PDI that rectify LLM estimates with human labels.

\section{Offensiveness is harder than politeness}

Overall, across models and prompting variants, coverage and $\Delta$ are worse on offensiveness than on politeness.  Several models that perform reasonably on politeness (e.g., llama-3-8b, mistral-large-2512) exhibit near-zero coverage and $\Delta > 30\%$ on offensiveness.  Two factors contribute: A lower base rate (32\% vs.\ 44\%) makes errors costlier on a percentage-point scale. The concept of ``offensive'' is more subjective and socially situated than ``polite,'' so models trained on different corpora disagree more with the human ground truth.

\section{Few-shot prompting is consistently best}

Few-shot prompting is the best or tied-best variant for 7 of 8 models on politeness and for 7 of 8 models on offensiveness (Table~\ref{tab:appendix_llm_eval}). The gains can be large: claude-haiku-4.5 improves from $\Delta = 15.5\%$ (zero-shot) to $6.0\%$ (few-shot) on politeness, and from 28.9\% to 5.3\% on offensiveness.

\section{Persona prompting often fails}\label{app:persona}

Persona prompting sometimes matches few-shot performance but can also hurt it, particularly on offensiveness: claude-haiku-4.5: 0\% coverage, $\Delta = 48.4\%$ (persona) vs. 54.5\% coverage, $5.3\%$ (few-shot); llama-3-8b: $\Delta = 64.3\%$ (persona) vs.\ $21.2\%$  (few-shot);  mistral-large-2512: $\Delta = 68.7\%$ (persona) vs.\ $34.1\%$ (few-shot). The persona instruction appears to shift predictions systematically in the wrong direction for these models, suggesting that persona-based framing is sensitive to how a model was trained.

\section{Models}

\paragraph{Politeness} Under the best prompting variant, models cluster into two tiers.
\emph{Strong performers} ($\Delta < 7\%$): gpt-oss-120b (5.3\%), mistral-large-2512 (5.7\%), gemini-3.1-pro (5.7\%), claude-haiku-4.5 (6.0\%), gpt-5.2 (6.4\%). \emph{Weaker performers} ($\Delta \geq 7\%$): llama-3-8b (9.5\%), claude-opus-4.6 (8.2\%), claude-sonnet-4.6 (6.6\% with persona but 10.5\% with few-shot).

\paragraph{Offensiveness}
\emph{Strong performers} (few-shot): gemini-3.1-pro ($\Delta = 5.0\%$),
claude-haiku-4.5 (5.3\%), claude-sonnet-4.6 (6.0\%), gpt-oss-120b (6.2\%). \emph{Weakest}: mistral-large-2512 (0\% coverage even with few-shot, $\Delta = 34.1\%$) and llama-3-8b (0\% coverage, $\Delta = 21.2\%$). These open-weight models appear poorly calibrated for detecting offensiveness.

\section{Robustness tests}\label{sec:sythetic}

\begin{figure*}[t]
    \centering
    \includegraphics[width=\textwidth]{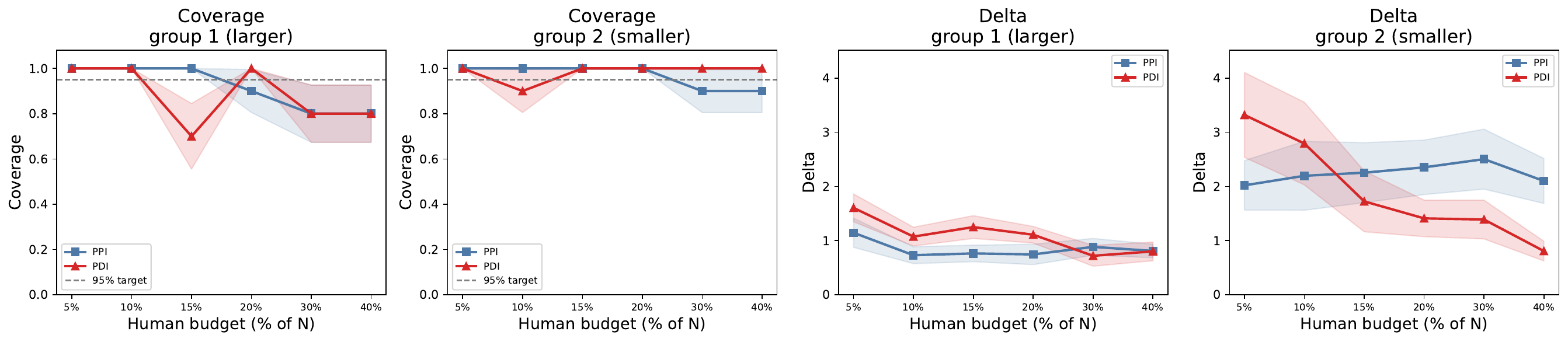}
    \caption{\textbf{Human annotation budget.} Coverage and delta of PPI and PDI as a function of human annotation budget (as a percentage of total sample size N), shown separately for group 1 (larger) and group 2 (smaller). Shaded regions indicate 95\% confidence bands across simulation runs.}
    \label{fig:synthetic1}
\end{figure*}

\begin{figure*}[t]
    \centering
    \includegraphics[width=\textwidth]{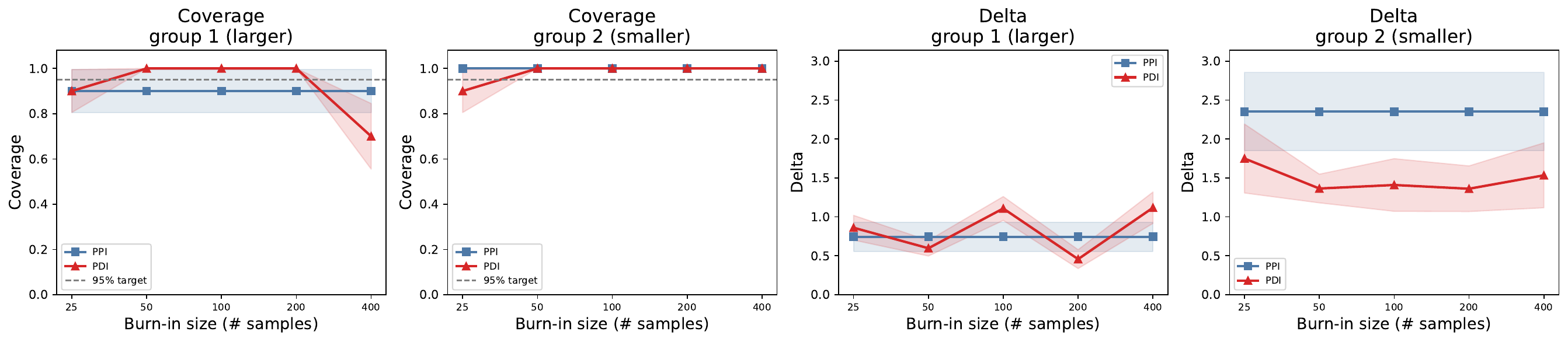}
    \caption{\textbf{Burn-in size.} Coverage and delta of PPI and PDI as a function of burn-in size, shown separately for group 1 (larger) and group 2 (smaller). Shaded regions indicate 95\% confidence bands across simulation runs.}
    \label{fig:synthetic2}
\end{figure*}

\begin{figure*}[t]
    \centering
    \includegraphics[width=\textwidth]{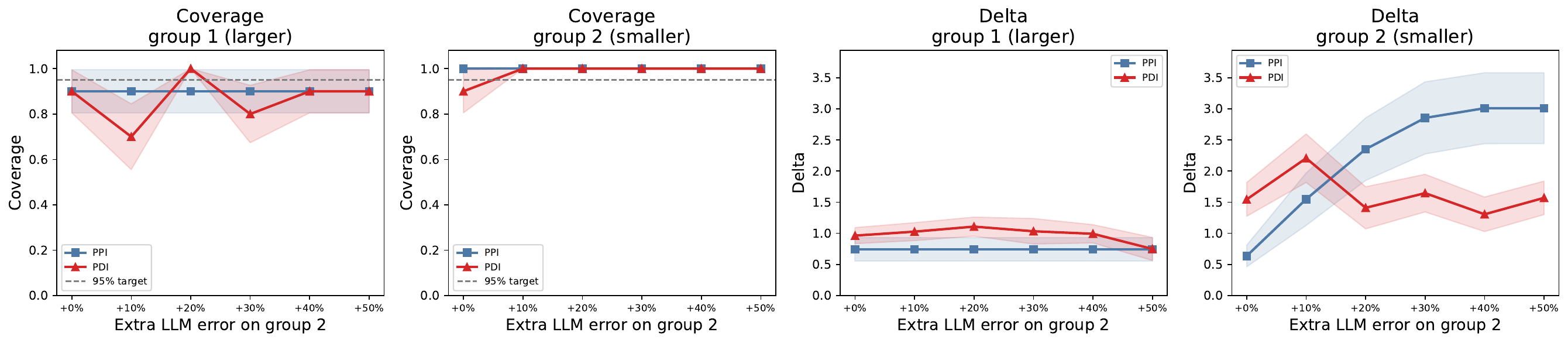}
    \caption{\textbf{LLM extra error on smaller group.} Coverage and delta of PPI and PDI as a function of LLM error discrepancy, shown separately for group 1 (larger) and group 2 (smaller).  Shaded regions indicate 95\% confidence bands across simulation runs.}
    \label{fig:synthetic3}
\end{figure*}

\begin{figure*}[t]
    \centering
    \includegraphics[width=\textwidth]{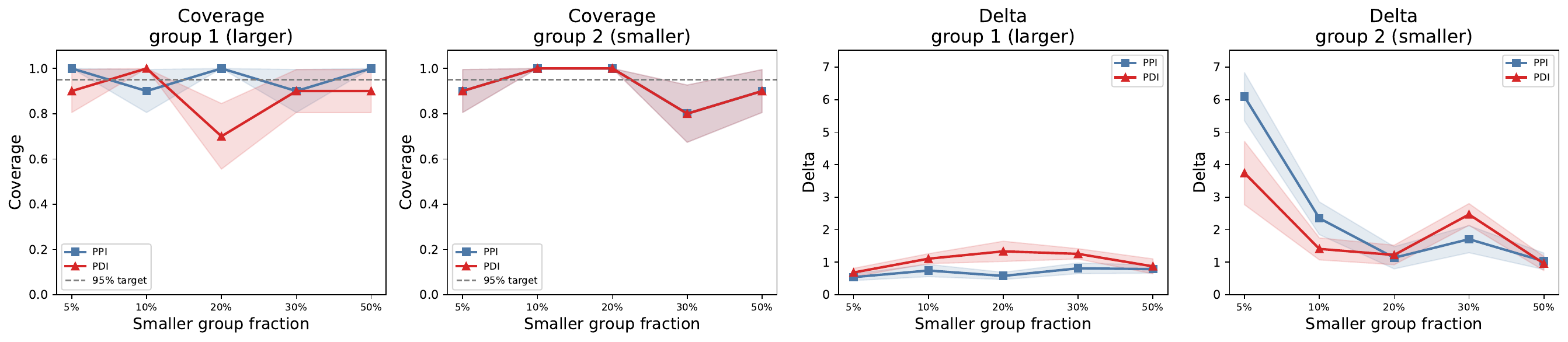}
    \caption{\textbf{Size of smaller group}. Coverage and delta of PPI and PDI as a function of the group skew, shown separately for group 1 (larger) and group 2 (smaller). Shaded regions indicate 95\% confidence bands across simulation runs.}
    \label{fig:synthetic4}
\end{figure*}

Figures~\ref{fig:synthetic1} to~\ref{fig:synthetic4} show robustness tests. Across four controlled experiments on synthetic data, PDI consistently outperforms PPI for the minority group while maintaining valid coverage for both methods. The precision advantage is largest when the human annotation budget is small, when the LLM makes more errors on the minority group, and when the minority group is small. These are the conditions where uniform sampling fails. Burn-in size has a modest effect on PDI's precision, with diminishing returns beyond 100 samples. On the majority group, PPI retains a slight edge throughout, since the LLM is already accurate there and uniform sampling suffices. PDI's adaptive allocation is most valuable when disparities in LLM performance across groups are large and groups are skewed.

\section{Disagreement analysis}\label{sec:new sec}

To identify the sources of annotation disagreement, we first applied the Fightin’ words method~\cite{Monroe_Colaresi_Quinn_2017} to uncover the words that are associated with the most disagreement between two groups of annotators. ($\geq$ 50 vs. other strata). We then applied LLooM~\cite{lam2024conceptInduction} to tokens with the highest human disagreement. Rather than analyzing individual words in isolation, this step helps us understand high-level themes that explain the disagreements.
\newline We used this method to: 
\begin{enumerate}
    \item Identify themes that annotators aged $\geq$ 50 years rated differently in terms of politeness than other strata, and 
    \item Identify themes that GPT-5.2 was differing in  annotations from human annotators. 
\end{enumerate}
The resulting themes are summarized in Tables~\ref{loom1} and~\ref{loom2}. In particular, annotators in the older stratum (>= 50 years) perceive differently texts that embed requests for information (e.g., ``if you have any questions, please feel free to call me''), hopeful sentiment (``look forward to seeing you tomorrow''), or polite closings.

\section{Additional datasets}
\label{app:additional-datasets}

To test whether our findings extend to other datasets with similar structure, we ran a supplementary analysis on three additional perspectivist datasets that release annotator-level labels together with annotator metadata:
\begin{itemize}
    \item \textbf{MultiPICo} \citep{casola-etal-2024-multipico}: irony detection for post--reply pairs from Reddit and Twitter. We use the English subset. Labels are binary (ironic vs.\ not ironic). The annotator attributes are gender, age, ethnicity, and nationality (UK, US, Ireland, Australia, India).
    \item \textbf{Conversational Sarcasm Corpus (CSC)} \citep{jang-frassinelli-2024-generalizable}: the sarcasm of a response given a described situation, rated by external evaluators on a 1--6 scale. We binarize ratings at $\geq 4$ (sarcastic). The evaluator attributes are gender, age, and ethnicity.
    \item \textbf{HS-Brexit} \citep{akhtar2021whose}: hate speech detection in tweets about Brexit, one of the LeWiDi datasets used by \citet{rizzi-etal-2024-soft}. Labels are binary. Each tweet is labeled by six annotators: a \emph{target group} of three Muslim immigrants in the UK and a \emph{control group} of three other annotators.
\end{itemize}

From each dataset we sample 1{,}000 (text, annotator) annotations. This matches the size of the POPQUORN subsets in the main experiments. Age is bucketed into 18--34, 35--49, and 50+, as in the main text. Annotators with missing gender or age, or with non-binary or undisclosed gender, are excluded, as in the main experiments.

We annotate every sampled instance with a single LLM, \texttt{gpt-oss-120b}, which we accessed through OpenRouter with default sampling parameters. We use the three prompting variants of Table~\ref{tab:prompts}, adapted to each task: zero-shot, few-shot, and persona. The few-shot variant uses 10 examples drawn at random from texts outside the sample. Persona prompts describe the annotator's demographic profile. For HS-Brexit, they describe the annotator as either a Muslim immigrant living in the UK or a UK resident who is not a Muslim immigrant.

\subsection{Exploratory analysis: does LLM performance vary across groups?}

\method{} can only improve over uniform sampling if the LLM's error depends on the annotator's group, because this dependence is what the error predictor learns and leverages. Before running PDI, we therefore test on each dataset whether LLM error varies across demographic groups. For each demographic dimension and prompting variant, we compute the per-group error rate: the fraction of annotations on which the binarized LLM label disagrees with the individual human label. We then test whether error is independent of group membership with a $\chi^2$ test. For reference, we also test whether the human labels themselves vary across groups. Groups with fewer than 30 annotations in the sample are excluded.

Table~\ref{tab:additional-exploratory} shows the results. Across all three datasets, the LLM's error rate is relatively flat across groups:
\begin{itemize}
    \item \textbf{MultiPICo:} per-group error rates lie between 0.25 and 0.39 for every dimension and variant.
    \item \textbf{CSC:} they lie between 0.23 and 0.39.
    \item \textbf{HS-Brexit:} they lie between 0.15 and 0.23.
\end{itemize}

None of the 21 tests on LLM error remains significant after Bonferroni correction ($\alpha = 0.05/21 \approx 0.002$). On MultiPICo and HS-Brexit, LLM error does not differ across groups. There, PDI's error predictor can only fit noise, so adaptive sampling reduces to uniform sampling (PPI) in expectation, with extra variance from the importance weights. We therefore do not evaluate PDI on these two datasets. 

We evaluate PDI on CSC only, since it has the largest error gap across groups: under few-shot prompting, which is the variant PPI and PDI use on CSC, error is 0.38 for Black evaluators and 0.23 to 0.30 for the other ethnicity groups (Table~\ref{tab:csc-results}). Of the three datasets, CSC is therefore the most relevant test case for error-driven upsampling.

\subsection{Setup on CSC}

We follow the protocol of main evaluation exactly:
\begin{itemize}
    \item total human annotation budget $n_{\text{human}} = 200$ out of $n = 999$ annotations (one instance whose zero-shot label could not be parsed is dropped);
    \item a uniform burn-in of $n_{\text{burnin}} = 100$, followed by $B = 3$ adaptive batches;
    \item smoothing with the uniform distribution at weight $\tau = 0.1$;
    \item an XGBoost error predictor trained on one-hot encoded evaluator gender, age, and ethnicity;
    \item 95\% confidence intervals, with results averaged over 20 trials.
\end{itemize}

The target $\theta^*_{g_k}$ is the share of annotations in group $g_k$ that rate the response as sarcastic ($\geq 4$). PPI and PDI use the prompting variant with the highest overall accuracy. This is few-shot (accuracy 0.708, vs.\ 0.705 for zero-shot and 0.683 for persona).

\subsection{Results}

Table~\ref{tab:csc-results} reports coverage and $\Delta$ for the ethnicity dimension, where group differences in LLM error are largest. Table~\ref{tab:csc-alloc} reports how many human annotations PDI and uniform PPI allocate to each group.

\paragraph{Coverage.} As on POPQUORN, the LLM-only baselines frequently miss the target. Averaged across ethnicity groups, coverage is 76.3\% for zero-shot, 41.3\% for few-shot, and 66.3\% for persona prompting. Few-shot, the most accurate variant overall, covers the White group in only 5\% of trials. PDI maintains coverage of at least 95\% for every ethnicity group (97.5\% on average). PPI's coverage falls to 85\% for Asian evaluators and 75\% for evaluators of mixed ethnicity.

\paragraph{Delta.} For Black evaluators, the group with the highest LLM error, PDI attains the lowest $\Delta$ of all methods (8.21, vs.\ 8.67 for PPI and at least 8.64 for the LLM-only baselines). Averaged across ethnicity groups, PPI has a slightly lower $\Delta$ than PDI (8.59 vs.\ 9.51). This mirrors the main results: PDI's benefit is targeted at the hardest group rather than uniform across groups.

\paragraph{Upsampling lift.} The error predictor successfully identifies Black evaluators as the high-error group. PDI allocates them 31.7 human annotations on average, compared with 22.9 under uniform PPI, a 38\% increase (Table~\ref{tab:csc-alloc}). For the other groups, PDI's allocation stays within $-13\%$ and $+8\%$ of uniform sampling, consistent with their near-identical error rates. Along the gender and age dimensions, where error rates are nearly flat (0.29--0.30 and 0.24--0.31), PDI and PPI both maintain coverage. PDI's coverage ranges from 85\% (Age 50+) to 100\%, and PPI has a lower $\Delta$ in every group, as expected when there is no error difference to exploit.

\subsection{Takeaway} On CSC, PDI behaves as it does on the POPQUORN tasks. The LLM-only baselines, including persona prompting, often fail to cover the group-level targets. PDI upsamples the group where the LLM is least accurate (Black evaluators, +38\% annotations) and achieves the lowest $\Delta$ and full coverage for that group. Averaged over groups, uniform PPI remains competitive or better on $\Delta$. The additional datasets also make the method's precondition concrete. PDI's gains depend on LLM error varying across annotator groups. On MultiPICo and HS-Brexit it does not, and there we expect PDI to offer no advantage over uniform PPI. Checking whether LLM error varies across groups, which requires only the burn-in sample, is therefore a useful first step before choosing between PDI and uniform sampling.

\newpage
\clearpage

\begin{table*}[t]
\centering
\small
\begin{tabular}{llcccc}
\toprule
\multirow{2}{*}{\textbf{Model}} & \multirow{2}{*}{\textbf{Prompting}} & \multicolumn{2}{c}{\textbf{Politeness}} & \multicolumn{2}{c}{\textbf{Offensiveness}} \\
\cmidrule(lr){3-4}\cmidrule(lr){5-6}
& & Cov. & $\Delta$  & Cov. & $\Delta$  \\
\midrule
  \texttt{gpt-5.2} & \textsc{zero-shot} & 0.36 & 6.5 & 0.09 & 18.7 \\
   & \textsc{few-shot} & 0.64 & 6.6 & 0.09 & 13.5 \\
   & \textsc{persona} & 0.55 & 6.4 & 0.09 & 18.4 \\
\addlinespace[3pt]
  \texttt{claude-sonnet-4.6} & \textsc{zero-shot} & 0.09 & 12.0 & 0.00 & 37.7 \\
   & \textsc{few-shot} & 0.27 & 10.5 & 0.64 & 6.0 \\
   & \textsc{persona} & 0.64 & 6.6 & 0.00 & 41.4 \\
\addlinespace[3pt]
  \texttt{claude-opus-4.6} & \textsc{zero-shot} & 0.45 & 8.2 & 0.45 & 8.0 \\
   & \textsc{few-shot} & 0.45 & 9.1 & 0.27 & 13.1 \\
   & \textsc{persona} & 0.45 & 8.3 & 0.45 & 8.2 \\
\addlinespace[3pt]
  \texttt{gemini-3.1-pro} & \textsc{zero-shot} & 0.55 & 5.7 & 0.27 & 9.4 \\
   & \textsc{few-shot} & 0.64 & 7.7 & 0.73 & 5.0 \\
   & \textsc{persona} & 0.55 & 6.3 & 0.18 & 14.3 \\
\addlinespace[3pt]
  \texttt{claude-haiku-4.5} & \textsc{zero-shot} & 0.00 & 15.5 & 0.00 & 28.9 \\
   & \textsc{few-shot} & 0.73 & 6.0 & 0.55 & 5.3 \\
   & \textsc{persona} & 0.09 & 14.5 & 0.00 & 48.4 \\
\addlinespace[3pt]
  \texttt{llama-3-8b} & \textsc{zero-shot} & 0.00 & 11.2 & 0.00 & 45.9 \\
   & \textsc{few-shot} & 0.27 & 9.5 & 0.09 & 21.2 \\
   & \textsc{persona} & 0.00 & 13.2 & 0.00 & 64.3 \\
\addlinespace[3pt]
  \texttt{mistral-large-2512} & \textsc{zero-shot} & 0.55 & 5.7 & 0.00 & 67.3 \\
   & \textsc{few-shot} & 0.64 & 6.4 & 0.00 & 34.1 \\
   & \textsc{persona} & 0.45 & 7.4 & 0.00 & 68.7 \\
\addlinespace[3pt]
  \texttt{gpt-oss-120b} & \textsc{zero-shot} & 0.55 & 6.6 & 0.18 & 15.1 \\
   & \textsc{few-shot} & 0.55 & 6.3 & 0.73 & 6.2 \\
   & \textsc{persona} & 0.55 & 5.3 & 0.27 & 13.1 \\
\bottomrule
\end{tabular}
\caption{LLM-only evaluation across 8 models and 3 prompting strategies on politeness and offensiveness annotation tasks. Cov.\ = coverage (higher is better). $\Delta$ = mean absolute error in percentage points (lower is better), averaged over all non-empty demographic subgroups.}
\label{tab:appendix_llm_eval}
\end{table*}

\newpage
\clearpage

\begin{table*}[hb!]
\centering
\small
\begin{tabular}{p{5cm} p{10cm}}
\toprule
\textbf{Prompt Type} & \textbf{Prompt Text} \\
\midrule

Politeness Zero Shot & Consider you read this email from a colleague. How polite is it? Answer using a 1--5 Likert scale where 1 means `not polite at all' and 5 means `very polite'. Return ONLY one integer. \\

Offensiveness Zero Shot & Consider you read this comment on Reddit. How offensive is it? Answer using a 1--5 Likert scale where 1 means `not offensive at all' and 5 means `very offensive'. Return ONLY one integer. \\

Politeness Few Shot & Here are example politeness ratings (1=very impolite, 5=very polite): \{Followed by Text: and Rating: for 10 examples. Rest of the prompt is the same as zero shot.\} \\

Offensiveness Few Shot & Here are example offensiveness ratings (1=not offensive, 5=extremely offensive): \{Followed by Text: and Rating: for 10 examples. Rest of the prompt is the same as zero shot.\} \\

Politeness Persona Prompt & You are \{persona\}. Answer all questions from the perspective of \{persona\}. Be consistent with this perspective, but still follow the instructions. \{followed by same prompt as Politeness Zero Shot\}. {persona} = a {gender} aged {age} with {education}\\

Offensiveness Persona Prompt & You are \{persona\}. Answer all questions from the perspective of \{persona\}. Be consistent with this perspective, but still follow the instructions. \{followed by same prompt as Offensiveness Zero Shot\}. \\
\end{tabular}
\caption{Prompts used for LLM annotation.}
\label{tab:prompts}
\end{table*}

\newpage
\clearpage

\begin{table*}[p]
\centering
\small
\renewcommand{\arraystretch}{1.2}
\caption{Example annotations illustrating inter-annotator disagreement on
offensiveness. Each row corresponds to one annotator.
A $\geq$50 annotator rating of 5.0 was required for selection.
\textbf{Bold} rating = annotator aged $\geq$50 who rated 5.0
(selection criterion).
{{\color{red!70!black}{Warning: content in this table may be upsetting or offensive.}}}
}
\label{tab:annotation-examples-off}

\begin{tabular}{p{0.7\textwidth}lc}
\toprule
\multicolumn{3}{c}{\textbf{Offensiveness} (1 = not offensive, 5 = very offensive)} \\
\midrule
\textbf{Text} & \textbf{Age} & \textbf{Rating} \\
\midrule

\multirow{9}{*}{\parbox{6.2cm}{\textit{``Women's studies is dumb though, stop defending your degree?''}}}
 & 18--24 & 5 \\
 & 25--29 & 5 \\
 & 30--34 & 3 \\
 & 40--44 & 5 \\
 & 45--49 & 1 \\
 & 50--54 & 1 \\
 & 50--54 & \textbf{5} \\
 & 54--59 & 1 \\
 & \multicolumn{2}{r}{\footnotesize mean = 3.25, std = 1.98} \\
\midrule

\multirow{11}{*}{\parbox{6.2cm}{\textit{``Still not as scary as autism.''}}}
 & 18--24 & 5 \\
 & 18--24 & 5 \\
 & 35--39 & 3 \\
 & 35--39 & 2 \\
 & 40--44 & 1 \\
 & 45--49 & 1 \\
 & 54--59 & \textbf{5} \\
 & 54--59 & 1 \\
 & $>$65  & 3 \\
 & $>$65  & 1 \\
 & \multicolumn{2}{r}{\footnotesize mean = 2.70, std = 1.77} \\
\midrule

\multirow{10}{*}{\parbox{6.2cm}{\textit{``Let's ban baths and haircuts for non-vaxxed people so everyone else knows to stay the hell away!''}}}
 & 18--24 & 5 \\
 & 25--29 & 2 \\
 & 30--34 & 1 \\
 & 40--44 & 1 \\
 & 45--49 & 4 \\
 & 54--59 & 1 \\
 & 60--64 & \textbf{5} \\
 & $>$65  & 1 \\
 & $>$65  & 5 \\
 & \multicolumn{2}{r}{\footnotesize mean = 2.78, std = 1.92} \\
\bottomrule
\end{tabular}
\end{table*}

\clearpage
\begin{table*}[p]
\centering
\small
\renewcommand{\arraystretch}{1.3}
\caption{Example annotations illustrating inter-annotator disagreement on
politeness. Each row corresponds to one annotator.
A $\geq$50 annotator rating of 5.0 was required for selection. \textbf{Bold} rating = annotator aged $\geq$50 who rated 5.0 (selection criterion). {{\color{red!70!black} {Warning: content in this table may be upsetting or offensive.}}}
}
\label{tab:annotation-examples-pol}

\begin{tabular}{p{0.7\textwidth}lc}
\toprule
\multicolumn{3}{c}{\textbf{Politeness} (1 = very impolite, 5 = very polite)} \\
\midrule
\textbf{Text} & \textbf{Age} & \textbf{Rating} \\
\midrule

\multirow{8}{*}{\parbox{6.2cm}{\textit{``No, tell her that it is off of Buffalo Speedway.''}}}
 & 18-24 & 1 \\
 & 25-29 & 3 \\
 & 30-34 & 1 \\
 & 30-34 & 5 \\
 & 50-54 & 4 \\
 & 60-64 & \textbf{5} \\
 & $>$65  & 1 \\
 & \multicolumn{2}{r}{\footnotesize mean = 2.86, std = 1.86} \\
\midrule

\multirow{7}{*}{\parbox{6.2cm}{\textit{``I would appreciate it if you didn't blame me for spills in your cube.''}}}
 & 18-24 & 1 \\
 & 18-24 & 1 \\
 & 18-24 & 1 \\
 & 30-34 & 4 \\
 & 54-59 & 3 \\
 & 60--64 & \textbf{5} \\
 & \multicolumn{2}{r}{\footnotesize mean = 2.50, std = 1.76} \\
\midrule

\multirow{8}{*}{\parbox{6.2cm}{\textit{``You can find my draft of the filing below if you are not already familiar with it.''}}}
 & 18-24 & 5 \\
 & 25-29 & 5 \\
 & 30-34 & 2 \\
 & 45-49 & 4 \\
 & 54-59 & 1 \\
 & 60-64 & \textbf{5} \\
 & $>$65  & 2 \\
 & \multicolumn{2}{r}{\footnotesize mean = 3.43, std = 1.72} \\
\bottomrule
\end{tabular}
\vspace{0em}
\end{table*}

\clearpage
\newpage

\begin{table*}[t]
\centering
\small
\setlength{\tabcolsep}{4pt}
\resizebox{\textwidth}{!}{%
\begin{tabular}{lrrp{5cm}p{4cm}}
\toprule
\textbf{Concept} & \textbf{Prevalence} & \textbf{Match Count} & \textbf{Meaning} & \textbf{Example }\\ 
\midrule
\multicolumn{5}{l}{} \\
\quad Request for Information   & 0.26 & 64 & Does the text include a request for information or clarification? & ``I HAVE NOT RECEIVED YOUR DATA FORMS… IF YOU HAVE ANY QUESTIONS, PLEASE FEEL FREE TO CALL ME. Thank you'' \\
\quad Hopeful Sentiment         & 0.18 & 45 & Does the text express hope or a positive outlook? & ``Look forward to seeing you tomorrow.'' \\
\quad Polite Closing            & 0.09 & 23 & Does the text include a polite closing or sign-off, such as 'best regards'? & ``I will leave the legal review up to your team. Please let me know if you have any questions about these CPs. Regards'' \\
\quad Attachment Mention        & 0.07 & 17 & Does the text mention an attachment or include the word 'attached'? & ``I am send you the following attached documents per request.'' \\
\quad Expression of Gratitude   & 0.04 & 9  & Does the text express gratitude or thanks? & ``I would appreciate if it you didn’t blame me for the spills in your cube'' \\
\bottomrule
\end{tabular}}
\caption{Themes with age-dependent differences ($\geq$ 50 vs. other strata). {{\color{red!70!black} {Warning: content in this table may be upsetting or offensive.}}}}
\label{loom1}
\end{table*}

\clearpage
\newpage

\begin{table*}[t]
\centering
\small
\setlength{\tabcolsep}{4pt}
\resizebox{\textwidth}{!}{%
\begin{tabular}{lrrp{5cm}p{4cm}}
\toprule
\textbf{Concept} & \textbf{Prevalence} & \textbf{Match Count} & \textbf{Meaning} & \textbf{Example } \\
\midrule
\multicolumn{5}{l}{} \\
\quad Request for Action        & 0.28 & 8 & Does the text contain a request for action or follow-up, indicated by words like 'please', 'let', or 'thanks'? & ``Please find attached the Summary and Hot List as of 11/16.  
Please contact me if you have any questions/comments.
Thanks'' \\
\quad Emotional Sharing         & 0.21 & 6 & Does the text involve sharing emotions or personal connections? & ``Are you in love with him?'' \\
\quad Evaluation and Quality    & 0.14 & 4 & Does the text involve evaluating the quality or value of something? & ``Let me tell you who won't win - RU.  They are arguably the worst
division 1 football team I've ever seen.  I saw them for the 1st time
this weekend on TV.  They are horrible.  Hope everything's going well.
Life at Enron sucks.  I'm sure you've been following the situation.
I'm sure there'll be a \#1 bestseller written about this circus.  Talk
to you soon.
KR'' \\
\quad Motivation and Support    & 0.10 & 3 & Does the text discuss the importance of encouragement or support? & ``Congrats on keeping your turf. I hope this question doesn't offend you, but would you trade it for an \$80 ENE price? Thanks.'' \\
\quad Travel Plans              & 0.07 & 2 & Does the text mention any travel plans or locations? & ``Hey give me a call! You should come to NYC this weekend to party with your friends!'' \\

\bottomrule
\end{tabular}}
\caption{Themes with Model-Human Disagreement (GPT-5.2 vs human annotators). {{\color{red!70!black} {Warning: content in this table may be upsetting or offensive.}}}}
\label{loom2}
\end{table*}

\newpage
\clearpage

\begin{table*}[t]
\centering
\small
\setlength{\tabcolsep}{4pt}
\resizebox{\textwidth}{!}{%
\begin{tabular}{llc cc cc cc cc}
\toprule
& & & \multicolumn{2}{c}{Human label} & \multicolumn{2}{c}{LLM error (zero shot)} & \multicolumn{2}{c}{LLM error (few shot)} & \multicolumn{2}{c}{LLM error (persona)} \\
\cmidrule(lr){4-5}\cmidrule(lr){6-7}\cmidrule(lr){8-9}\cmidrule(lr){10-11}
Dataset & Dimension & $K$ & Range & $p$ & Range & $p$ & Range & $p$ & Range & $p$ \\
\midrule
MultiPICo & Gender      & 2 & 0.30--0.34 & 0.272 & 0.29--0.32 & 0.299 & 0.30--0.33 & 0.391 & 0.29--0.32 & 0.318 \\
          & Age         & 3 & 0.29--0.36 & 0.249 & 0.30--0.35 & 0.398 & 0.28--0.39 & 0.044 & 0.27--0.35 & 0.180 \\
          & Ethnicity   & 4 & 0.25--0.39 & 0.135 & 0.26--0.35 & 0.546 & 0.31--0.36 & 0.752 & 0.25--0.36 & 0.223 \\
          & Nationality & 5 & 0.28--0.37 & 0.406 & 0.27--0.36 & 0.352 & 0.28--0.36 & 0.517 & 0.25--0.38 & 0.031 \\
\midrule
CSC       & Gender      & 2 & 0.31--0.35 & 0.140 & 0.28--0.31 & 0.323 & 0.29--0.30 & 0.839 & 0.29--0.35 & 0.083 \\
          & Age         & 3 & 0.26--0.36 & 0.048 & 0.27--0.31 & 0.363 & 0.24--0.31 & 0.229 & 0.30--0.33 & 0.689 \\
          & Ethnicity   & 4 & 0.28--0.41 & 0.231 & 0.29--0.37 & 0.477 & 0.23--0.38 & 0.161 & 0.27--0.39 & 0.457 \\
\midrule
HS-Brexit & Annotator group & 2 & 0.05--0.21 & $<$0.001 & 0.17--0.21 & 0.115 & 0.15--0.17 & 0.614 & 0.19--0.23 & 0.138 \\
\bottomrule
\end{tabular}}
\caption{\textbf{Exploratory analysis on additional datasets.} For each demographic dimension ($K$ groups with $\geq 30$ annotations in the 1{,}000-annotation sample), we show the range of per-group rates across groups and the $p$-value of a $\chi^2$ test of independence between group and the quantity. \emph{Human label} is the per-group positive rate. \emph{LLM error} is the per-group rate at which the binarized \texttt{gpt-oss-120b} label disagrees with the human label. No test of LLM error survives Bonferroni correction across the 21 tests.}
\label{tab:additional-exploratory}
\end{table*}

\begin{table*}[t]
\centering
\small
\setlength{\tabcolsep}{4.5pt}
\begin{tabular}{l cc cc cc cc cc}
\toprule
& \multicolumn{2}{c}{Avg (Ethnicity)} & \multicolumn{2}{c}{White} & \multicolumn{2}{c}{Black} & \multicolumn{2}{c}{Asian} & \multicolumn{2}{c}{Mixed} \\
\cmidrule(lr){2-3}\cmidrule(lr){4-5}\cmidrule(lr){6-7}\cmidrule(lr){8-9}\cmidrule(lr){10-11}
Method & Cov. & $\Delta$ & Cov. & $\Delta$ & Cov. & $\Delta$ & Cov. & $\Delta$ & Cov. & $\Delta$ \\
\midrule
\multicolumn{11}{l}{\textit{Group size $n$ / LLM error (few shot)}} \\
& \multicolumn{2}{c}{999 / 0.29} & \multicolumn{2}{c}{746 / 0.28} & \multicolumn{2}{c}{117 / 0.38} & \multicolumn{2}{c}{70 / 0.30} & \multicolumn{2}{c}{43 / 0.23} \\
\midrule
gpt-oss-120b (zero shot) & \low{76.3} & 11.63 & \low{35.0} & 8.86  & 95.0        & 8.64  & \low{80.0} & 14.73 & 95.0       & 14.28 \\
gpt-oss-120b (few shot)  & \low{41.3} & 17.16 & \low{5.0}  & 12.82 & \low{35.0}  & 20.22 & \low{45.0} & 19.90 & \low{80.0} & 15.69 \\
gpt-oss-120b (persona)   & \low{66.3} & 13.43 & \low{20.0} & 11.52 & 90.0        & 8.84  & \low{75.0} & 13.90 & \low{80.0} & 19.46 \\
PPI                      & \low{86.3} & 8.59  & 90.0       & 2.62  & 95.0        & 8.67  & \low{85.0} & 8.82  & \low{75.0} & 14.25 \\
PDI                      & 97.5       & 9.51  & 100.0      & 3.14  & {100.0} & {8.21} & 95.0 & 12.44 & 95.0     & 14.26 \\
\bottomrule
\end{tabular}
\caption{\textbf{CSC (sarcasm).} Coverage (Cov., \%) and delta ($\Delta$, percentage points) by evaluator ethnicity (\texttt{gpt-oss-120b}, $n_{\text{human}} = 200$, 20 trials). Values below 90\% coverage are highlighted. Bold indicates the best performance for Black evaluators, the group with the highest LLM error. PPI and PDI use the few-shot variant. The \emph{Avg} group-size cell shows the total number of annotations and the overall few-shot error rate.}
\label{tab:csc-results}
\end{table*}

\begin{table*}[t]
\centering
\small
\setlength{\tabcolsep}{4pt}
\begin{tabular}{llccc}
\toprule
Dimension & Group & LLM err. & PDI & PPI \\
\midrule
Gender    & Woman     & 0.30 & 110.8 & 105.2 \\
          & Man       & 0.29 & 89.2  & 94.8  \\
\midrule
Age       & Age 18--34 & 0.31 & 95.6  & 96.8  \\
          & Age 35--49 & 0.30 & 75.2  & 69.8  \\
          & Age 50+    & 0.24 & 29.2  & 33.5  \\
\midrule
Ethnicity & White     & 0.28 & 141.5 & 149.9 \\
          & Black     & 0.38 & {31.7}  & 22.9  \\
          & Asian     & 0.30 & 13.0  & 13.9  \\
          & Mixed     & 0.23 & 8.3   & 9.1   \\
\bottomrule
\end{tabular}
\caption{\textbf{CSC.} Average number of human annotations allocated per group by PDI and PPI, averaged across 20 trials (total budget $200$ per trial). \emph{LLM err.} is the per-group error rate of the few-shot variant. PDI allocates 38\% more annotations to Black evaluators than PPI (31.7 vs.\ 22.9), the group with the highest LLM error.}
\label{tab:csc-alloc}
\end{table*}

\end{document}